\RequirePackage{fix-cm}
\documentclass[twocolumn]{svjour3}

\smartqed  

\usepackage{amsmath}
\usepackage{amssymb}
\usepackage{graphicx}
\usepackage{hyperref}
\usepackage{titlesec}
\usepackage{subcaption}
\usepackage{placeins}
\usepackage{amsmath}
\usepackage{graphicx}
\usepackage{url}
\usepackage{booktabs}
\usepackage{natbib}
\usepackage{orcidlink}

\titlespacing{\section}{0pt}{12pt plus 4pt minus 2pt}{6pt plus 2pt minus 2pt}

\begin{document}

\title{Machine Learning Techniques with Fairness for Prediction of Completion of Drug and Alcohol Rehabilitation}


\author{Karen Roberts-Licklider         \and
        Theodore Trafalis
}


\institute{K. Roberts-Licklider \at
              University of Oklahoma
              School of Industrial and Systems Engineering\\
              \email{Karen.R.RobertsLicklider-1@ou.edu}  
           \and
           T. Trafalis \at
              University of Oklahoma
              School of Industrial and Systems Engineering
              \email{ttrafalis@ou.edu}
}

\date{Received: date / Accepted: date}

\maketitle

\begin{abstract}
The aim of this study is to look at predicting whether a person will complete a drug and alcohol rehabilitation program and the number of times a person attends.  The study is based on demographic data obtained from Substance Abuse and Mental Health Services Administration (SAMHSA) from both admissions and discharge data from drug and alcohol rehabilitation centers in Oklahoma.  Demographic data is highly categorical which led to binary encoding being used and various fairness measures being utilized to mitigate bias of nine demographic variables.  Kernel methods such as linear, polynomial, sigmoid, and radial basis functions were compared using support vector machines at various parameter ranges to find the optimal values.  These were then compared to methods such as decision trees, random forests, and neural networks. Synthetic Minority Oversampling Technique Nominal (SMOTEN) for categorical data was used to balance the data with imputation for missing data. The nine bias variables were then intersectionalized to mitigate bias and the dual and triple interactions were integrated to use the probabilities to look at worst case ratio fairness mitigation.  Disparate Impact, Statistical Parity difference, Conditional Statistical Parity Ratio, Demographic Parity, Demographic Parity Ratio, Equalized Odds, Equalized Odds Ratio, Equal Opportunity, and Equalized Opportunity Ratio were all explored at both the binary and multiclass scenarios.  
\end{abstract}

\section{Introduction}
\label{intro}
Substance abuse is one of the leading causes for mental illness and these issues are dealt with in the Diagnostic and Statistical Manual of Mental Disorders (DSM) and the International Classification of Diseases (ICD) \cite{dsm5}. 36.2\% of adults age 18-25 had a mental illness, 29.4\% for age 26-49 and 13.9\% 50 or older with 11.6\% being serious mental illness for age 18-25, 7.6\% for 26-49 years old and 3.0\% for 50 years or older \cite{nsduh2022}. The COMPAS assessment has been used in criminal sentencing to predict whether an offender will reoffend, however it has proven to be racially biased. COMPAS is the Correctional Offender Management Profiling for Alternative Sanctions. It was developed in 1998 and uses a recidivism risk scale and has been used since 2000. It predicts a defendant’s risk of committing a misdemeanor or felony within 2 years of assessment for 137 features about an individual and the individual’s past criminal record \cite{dressel2018accuracy}. The COMPAS assessment incorrectly predicted that whites would reoffend at a rate of 47.7\% which was twice the rate of blacks at 28.0\%. It favored white defendants over blacks. Its accuracy for white defendants was 67\% and 63.8\% for black defendants \cite{dressel2018accuracy}. For this reason, considering fairness in these types of assessments is important. Also looking at whether an offender would complete rehab instead of going to prison is important as an alternative to sending them to prison due to overcrowding in prisons especially in Oklahoma. Oklahoma itself ranks 3rd if looking at the world’s incarceration rate considering every state as a country \cite{wildra2021}. For the scope of this paper, treatment episode data from the Substance Abuse and Mental Health Services website \cite{samsa2021} is used to look at predicting whether a person completed a drug and alcohol rehabilitation program, the number of prior treatments a person has been to when entering treatment, and then the concatenation of both variables.\cite{chawla2002} thoroughly reviewed \\ SMOTE: Synthetic  Minority Over-Sampling Technique for both the continuous and Nominal case. The nominal case is what we will use from this work. The algorithms and use of the value distance metric were described for SMOTE-N and SMOTE-NC and how these relate to the ROC curve. Dual and three-way interactions of both the additive and multiplicative type are discussed in \cite{veenstra2011} and the intersecting of bias terms in shown in this work. Demographic Parity, Disparate Impact, Statistical Parity Difference, Equal Opportunity, Equalized Odds, and Min-Max Fairness were all discussed in length in \cite{irfan2023}. They all compared models such as Support Vector Classifiers, Gaussian Process Classifiers, Gaussian Naïve Bayesian, and Linear Discriminant Analysis. Worst-case fairness metrics such as Demographic Parity Ratio, Disparate Impact Ratio, Conditional Statistical Parity Ratio, Equal Opportunity Ratio, and Equal Odds Ratio as well as a multiclass example was shown in \cite{ghosh2021}. These fairness measures can be used when there are multiple classes or when the reweighting or intersectionality has been done and there are more than two ratios to evaluate between.
This paper is organized as follows: in section 2 we discuss the methodology. In sections 3, 4 and 5 we discuss issues of data cleaning, encoding, balancing and data sensitivities and distributions. Section 6 discusses fairness measures applied to our data. In sections 7 and 8 we explore several machine learning models and interpretation of the results. Sections 9 and 10 discusses reweighting and new fairness calculations. Finally sections 11 and 12 discuss the conclusions and future work.

\section{Methodology}
Various machine learning algorithms were used on this data with a focus on kernel methods for support vector machines. With this data being highly categorical in nature, encoding techniques were used to transform the data to make it more manageable.  The data was balanced to make predictions more accurate, and the missing data was imputed.  Fairness measures were compared before and after weighing the variables using two- and three-way interactions with chi squared significance and intersectionalizing the nine bias variables.  The nine bias variables explored were gender, veteran status, marital status, education, age, employment, pregnancy status, race, and ethnicity.  Four kernels were compared; linear, polynomial, radial basis function, and sigmoid.  These were tried at a range of degrees, c values, gammas, and r values. Decision trees, random forests, and neural networks were compared. The models were then compared to each other to see which models outperformed each other.  

\section{Data Clean Up Process}
The data set was filtered down to Oklahoma and down to 30-day rehabilitation centers only.   A calculated column was created called COMPLETED in which a value of ‘COMPLETE’ was taken if the REASON variable took on a value of 1 and ‘INCOMPLETE’ if REASON took on a value of anything else.  The reason code translations can be seen in figure 1 below.  

\begin{figure}[ht!]
  \includegraphics[width=0.45\textwidth]{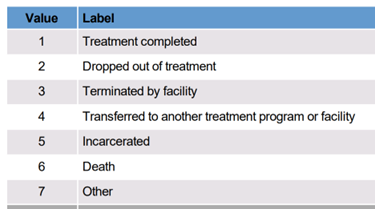}
\caption{Reason Codes}
\label{fig:1}       
\end{figure}

Some user defined functions were created to clean up the data. One removes unneeded columns that just have a constant value in them or had an identification number for the CASEID, admit year (ADMYR), REGION and STFIPS could be removed since the data was filtered for Oklahoma.  STFIPS was the FIPS code for each state.  This was filtered for 40 for Oklahoma as mentioned above.

\section{Encoding and Balancing Data}
Next, three data sets were created.  One was for the completed predictions, one was for the reasons predictions, and one was for the no-priors (number of times in treatment) predictions.  Each data set went through an encoding and data balancing processes.  One hot encoding was used to encode the categorical variables into new binary variables.  Each class within that variable is created into a new column which takes on the value of 1 if it appears for that record and 0 otherwise. An Example of how the services classes in figure 2 are translated to binary variables in figure 3 can be seen below. A description of what each service class is, is included.
\begin{figure}[ht!]
  \includegraphics[width=0.50\textwidth]{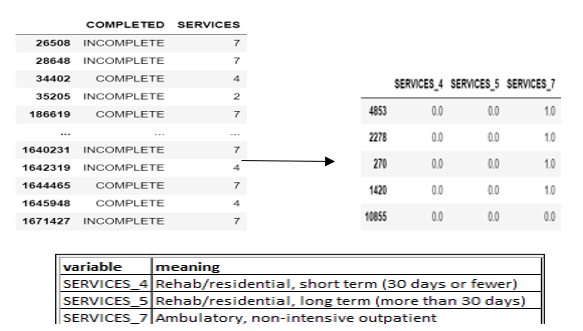}
\caption{Services One Hot Encoding}
\label{fig:2}       
\end{figure}

Next, we look at how the data is imbalanced in all three data sets.  We use the SMOTEN (Synthetic Minority Over-Sampling Technique Nominal) package in Python which uses K-Nearest neighbors \cite{gajawada} algorithm to balance the data.  See the figures below. This is done for the three data frames used in prediction.
\begin{figure}[ht!]
  \includegraphics[width=0.50\textwidth]{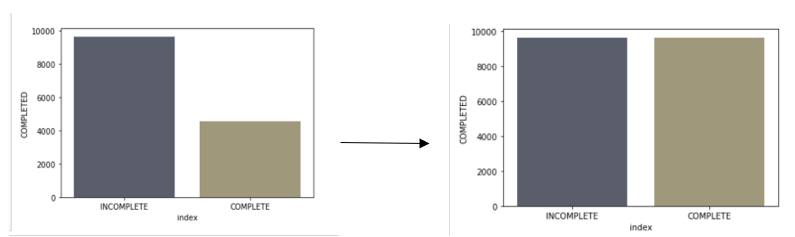}
\caption{Completed SMOTE Applied}
\label{fig:3}       
\end{figure}

\begin{figure}[ht!]
  \includegraphics[width=0.50\textwidth]{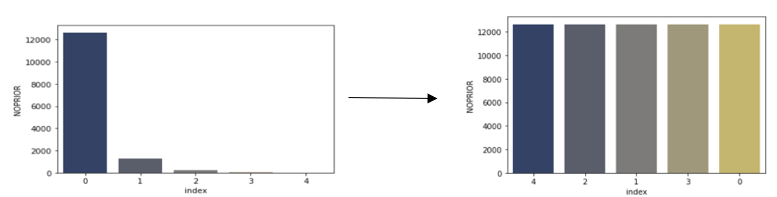}
\caption{NOPRIOR SMOTE Applied}
\label{fig:4}       
\end{figure}

\begin{figure}[ht!]
  \includegraphics[width=0.50\textwidth]{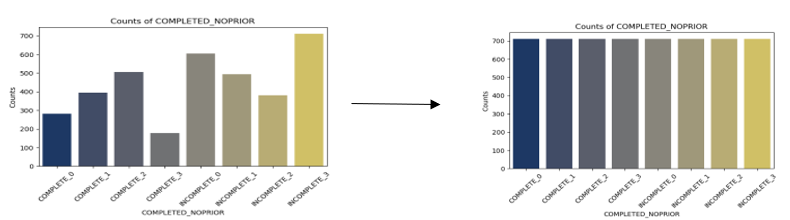}
\caption{Completed\_NORPIOR SMOTE Applied}
\label{fig:5}       
\end{figure}

Typically SMOTE is used which helps to improve the prediction accuracy. It distributes the instances of the majority class and the minority class equally. SMOTE technique increases the predictive accuracy over the minority class by creating synthetic instances of that minority class, \cite{jishan2015}. SMOTEN extends SMOTE for nominal features by getting the nearest neighbors by using the modified version of the Value Difference Metric which looks at the overlap of feature values over all feature vectors.  A matrix of features for all feature vectors is created and the distance between those features is defined in the following equation:

\begin{equation}
\delta\left(V_1,V_2\right) = \sum_{i=1}^{n} \left| \frac{C_{1i}}{C_1} - \frac{C_{2i}}{C_2} \right|^k
\end{equation}

Where \( V_1 \) and \( V_2 \) are the corresponding feature values and \( C_1 \) is the total number of times \( V_1 \) occurs and \( C_{1i} \) is the total occurrences of feature \( V_1 \) for class \( i \). The same holds for \( V_2 \), \( C_2 \), and \( C_{2i} \). \( k \) is some constant that is typically set to 1. This gives the matrix of value differences for each nominal value in the set of feature vectors\cite{chawla2002}. An example given can be seen on the next page.

\begin{figure}[ht!]
  \includegraphics[width=0.45\textwidth]{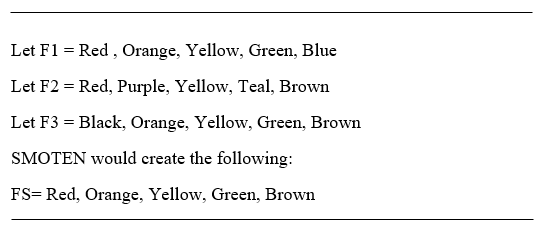}
\caption{SMOTEN Example}
\label{fig:6}       
\end{figure}

\section{Data Sensitivities/Distributions}
\subsection{Pareto Charts}
Nine variables were chosen and distributed by Pareto charts to see where bias occurred. These variables were gender, race, age, ethnic, veteran status, education status, marital status, employment status, and pregnancy status. These are shown for completion rehab outcomes in the figures below.

\begin{figure}[ht!]
\centering
\begin{subfigure}[b]{0.45\textwidth}
   \includegraphics[width=\linewidth]{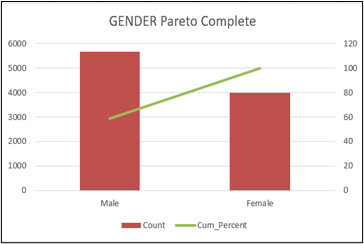}
   \caption{Gender Pareto}
   \label{fig:7}
\end{subfigure}
\hfill
\begin{subfigure}[b]{0.45\textwidth}
   \includegraphics[width=\linewidth]{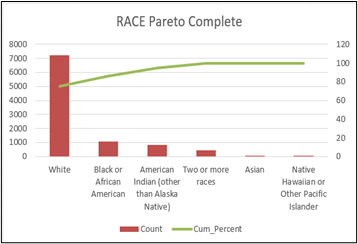}
   \caption{Race Pareto}
   \label{fig:8}
\end{subfigure}
\hfill
\begin{subfigure}[b]{0.45\textwidth}
   \includegraphics[width=\linewidth]{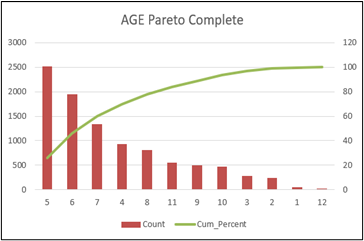}
   \caption{Age Pareto}
   \label{fig:9}
\end{subfigure}
\caption{Pareto Charts for Gender, Race, and Age}
\end{figure}

\begin{figure}[ht!]
\centering
\begin{subfigure}[b]{0.45\textwidth}
   \includegraphics[width=\linewidth]{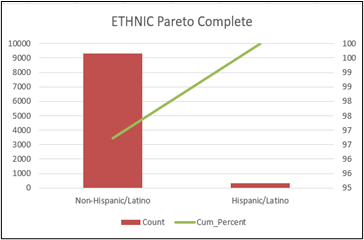}
   \caption{ETHNIC Pareto}
   \label{fig:10}
\end{subfigure}
\hfill
\begin{subfigure}[b]{0.45\textwidth}
   \includegraphics[width=\linewidth]{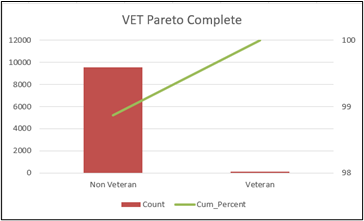}
   \caption{VET Pareto}
   \label{fig:11}
\end{subfigure}
\hfill
\begin{subfigure}[b]{0.45\textwidth}
   \includegraphics[width=\linewidth]{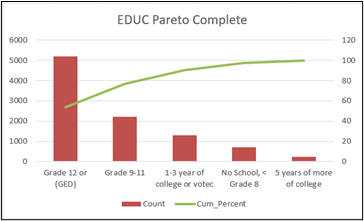}
   \caption{EDUC Pareto}
   \label{fig:12}
\end{subfigure}
\caption{Pareto Charts for Ethnicity, Veteran Status, and Education Status}
\end{figure}

\begin{figure}[ht!]
\centering
\begin{subfigure}[b]{0.45\textwidth}
   \includegraphics[width=\linewidth]{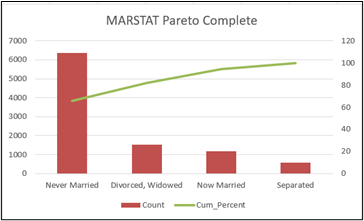}
   \caption{MARSTAT Pareto}
   \label{fig:13}
\end{subfigure}
\hfill
\begin{subfigure}[b]{0.45\textwidth}
   \includegraphics[width=\linewidth]{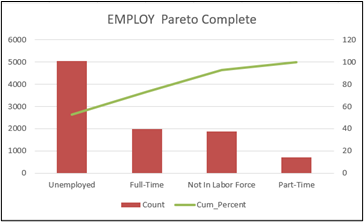}
   \caption{EMPLOY Pareto}
   \label{fig:14}
\end{subfigure}
\hfill
\begin{subfigure}[b]{0.45\textwidth}
   \includegraphics[width=\linewidth]{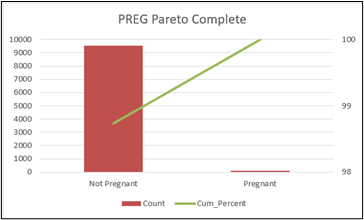}
   \caption{PREG Pareto}
   \label{fig:15}
\end{subfigure}
\caption{Pareto Charts for Marital Status, Employment Status, and Pregnancy Status}
\end{figure}
\FloatBarrier
\subsection{Bucketized Categories}
Based on the Pareto charts above, each variable was bucketized into dichotomous variables.  Variables such as gender, pregnancy, ethnicity, and veteran status were already dichotomous so they did not change.  Race became either white or non-white, employment status became employed or not employed, age became under 40 or 40 plus, marital status became never married or married/previously married, and education status became college or no college.
\section{Fairness Measures}
\subsection{Disparate Impact}
To find where the discrimination occurs, we first look at disparate impact in the dichotomous variables.  
\begin{equation}
    DI = \frac{P(\hat{Y}=1 \mid A=0)}{P(\hat{Y}=1 \mid A=1)}
\end{equation}

This compares the proportion of individuals receiving a favorable outcome for a privileged and underprivileged group. The closer to 1 it is, the more fair it is. \( \hat{Y} \) is the model predictions and \( A \) is the protected attribute, with 0 being the underprivileged class and 1 being the privileged class \cite{irfan2023}. The resulting disparate impact charts for the completed model can be seen in the figures below. Typically the 80\% rule is followed for the threshold to be considered discriminatory \cite{ghosh2021}.

\begin{figure}
\centering
\begin{subfigure}[b]{0.45\textwidth}
   \includegraphics[width=\linewidth]{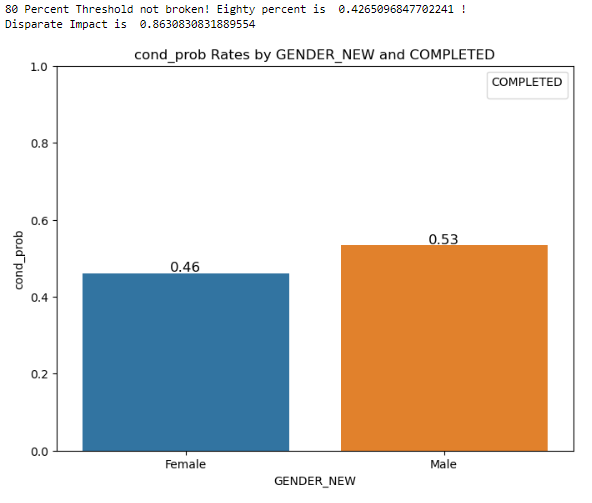}
   \caption{DI Dichotomous Gender}
   \label{fig:10a}
\end{subfigure}
\hfill
\begin{subfigure}[b]{0.45\textwidth}
   \includegraphics[width=\linewidth]{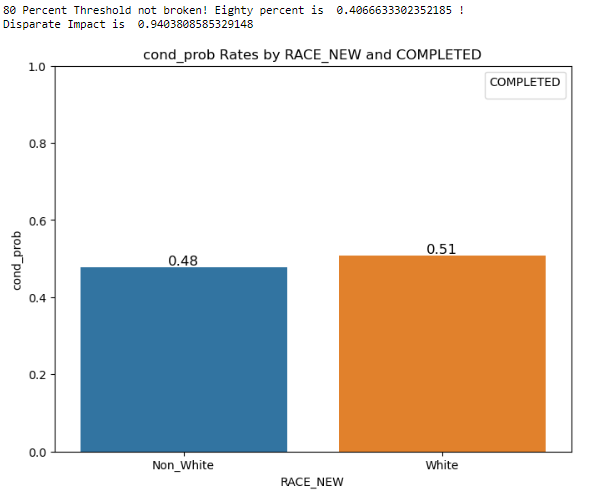}
   \caption{DI Dichotomous Race}
   \label{fig:10b}
\end{subfigure}
\hfill
\begin{subfigure}[b]{0.45\textwidth}
   \includegraphics[width=\linewidth]{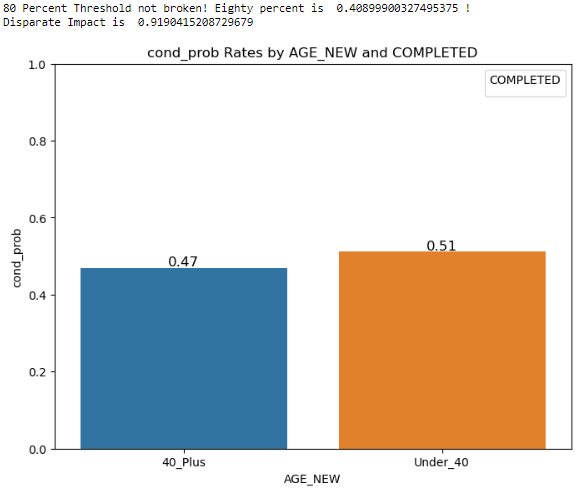}
   \caption{DI Dichotomous Age}
   \label{fig:10c}
\end{subfigure}
\caption{DI Dichotomous Charts for Gender, Race, and Age}
\end{figure}

\begin{figure}[h]
\centering
\begin{subfigure}[b]{0.45\textwidth}
   \includegraphics[width=\linewidth]{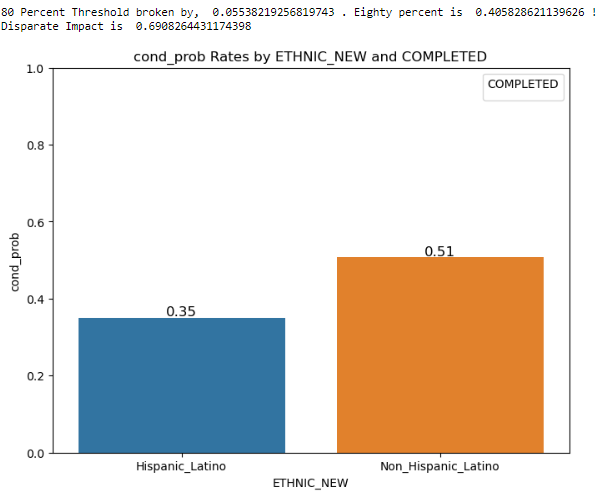}
   \caption{DI Dichotomous ETHNIC}
   \label{fig:11a}
\end{subfigure}
\hfill
\begin{subfigure}[b]{0.45\textwidth}
   \includegraphics[width=\linewidth]{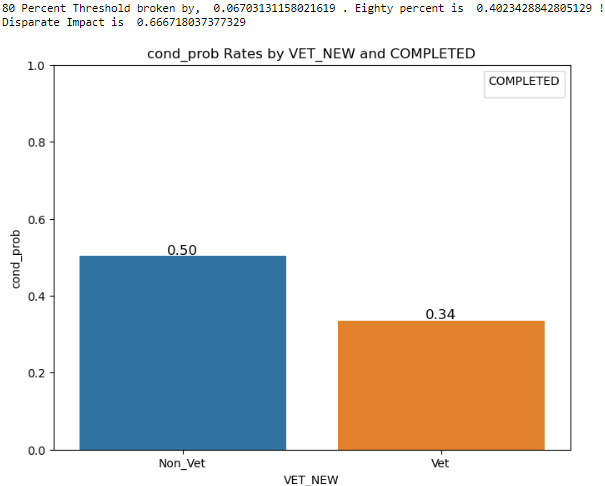}
   \caption{DI Dichotomous VET}
   \label{fig:11b}
\end{subfigure}
\hfill
\begin{subfigure}[b]{0.45\textwidth}
   \includegraphics[width=\linewidth]{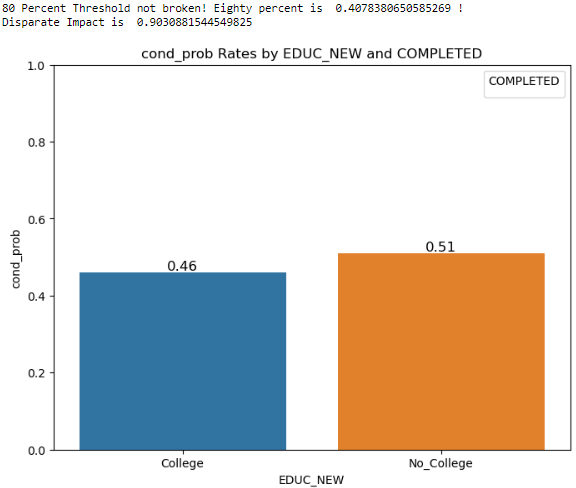}
   \caption{DI Dichotomous EDUC}
   \label{fig:11c}
\end{subfigure}
\caption{DI Dichotomous Charts for Ethnicity, Veteran Status, and Education Status}
\end{figure}

\begin{figure}[h]
\centering
\begin{subfigure}[b]{0.45\textwidth}
   \includegraphics[width=\linewidth]{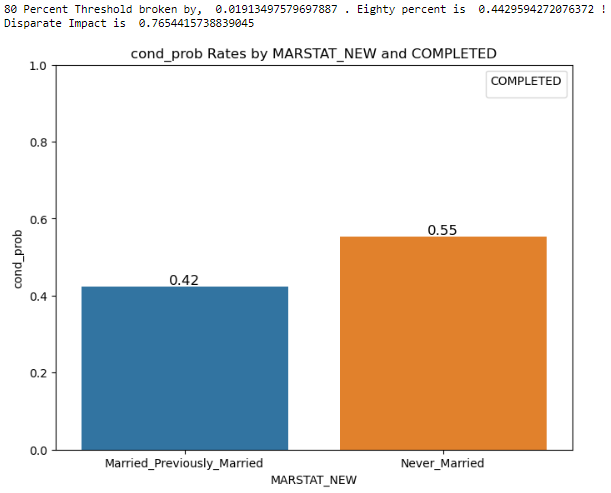}
   \caption{DI Dichotomous MARSTAT Pareto}
   \label{fig:12a}
\end{subfigure}
\hfill
\begin{subfigure}[b]{0.45\textwidth}
   \includegraphics[width=\linewidth]{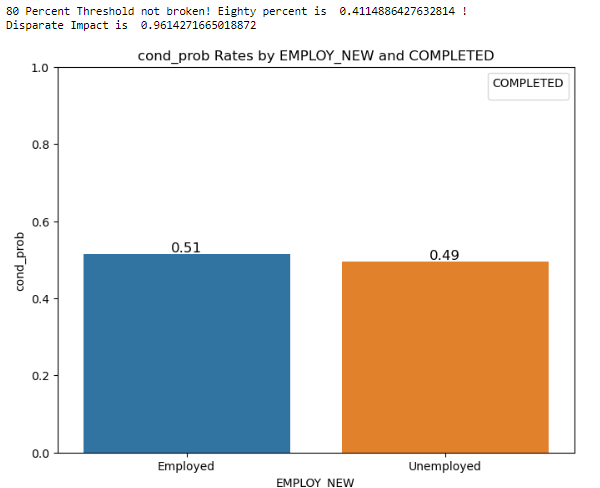}
   \caption{DI Dichotomous EMPLOY}
   \label{fig:12b}
\end{subfigure}
\hfill
\begin{subfigure}[b]{0.45\textwidth}
   \includegraphics[width=\linewidth]{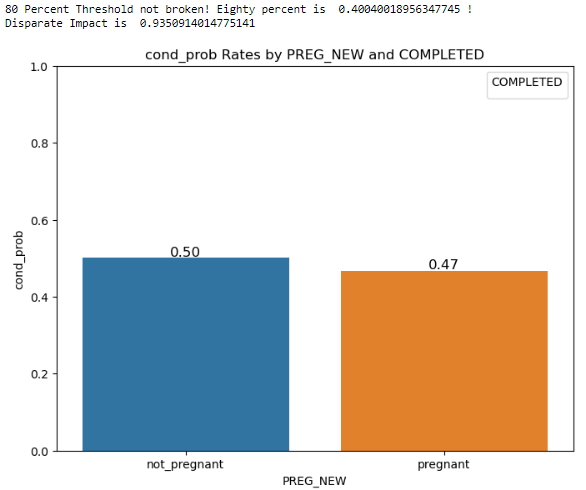}
   \caption{DI Dichotomous PREG}
   \label{fig:12c}
\end{subfigure}
\caption{DI Dichotomous Charts for Marital Status, Employment Status, and Pregnancy Status}
\end{figure}

\FloatBarrier

\subsection{Disparate Impact – Multiclass}
Next we look at the multiclass case in which we find the disparate impact for each class and divide the minimum conditional probability when the target matches the outcome by the maximum conditional probability when the target matches the outcome.    

\begin{equation}
    \mathrm{DI}_{\text{Multiclass}} = \frac{\min \left( P(\hat{Y} = 1 \mid A = 1) \right)}{\max \left( P(\hat{Y} = 1 \mid A = 1) \right)}
\end{equation}

The outputs for noprior model for each of the variables can be seen in the table below.  

\begin{table}[h]
\centering
\caption{Disparate Impact Analysis}
\label{tab:my_label}
\begin{tabular}{llll}
\hline
Variable & Class & DI   & 80\% Threshold Broken \\ \hline
GENDER   & 0     & 0.43 & UNFAIR               \\
         & 3     & 0.78 & UNFAIR               \\
         & 2     & 0.02 & UNFAIR               \\
         & 1     & 0.81 & FAIR                 \\
AGE      & 0     & 0.56 & UNFAIR               \\
         & 3     & 0.01 & UNFAIR               \\
         & 2     & 0.65 & UNFAIR               \\
         & 1     & 0.97 & FAIR                 \\
VET      & 0     & 0.28 & UNFAIR               \\
         & 3     & 1.00 & FAIR                 \\
         & 2     & 0.17 & UNFAIR               \\
         & 1     & 0.33 & UNFAIR               \\
EDUC     & 0     & 0.80 & UNFAIR               \\
         & 3     & 1.00 & FAIR                 \\
         & 2     & 0.38 & UNFAIR               \\
         & 1     & 0.73 & UNFAIR               \\
MARSTAT  & 0     & 0.58 & UNFAIR               \\
         & 3     & 0.42 & UNFAIR               \\
         & 2     & 0.67 & UNFAIR               \\
         & 1     & 0.77 & UNFAIR               \\
EMPLOY   & 0     & 0.36 & UNFAIR               \\
         & 3     & 1.00 & FAIR                 \\
         & 2     & 1.00 & FAIR                 \\
         & 1     & 0.75 & UNFAIR               \\
RACE     & 0     & 0.92 & FAIR                 \\
         & 3     & 0.00 & UNFAIR               \\
         & 2     & 0.34 & UNFAIR               \\
         & 1     & 0.81 & FAIR                 \\
ETHNIC   & 0     & 0.64 & UNFAIR               \\
         & 3     & 1.00 & FAIR                 \\
         & 2     & 0.43 & UNFAIR               \\
         & 1     & 0.20 & UNFAIR               \\
PREG     & 0     & 0.30 & UNFAIR               \\
         & 3     & 0.17 & UNFAIR               \\
         & 2     & 1.00 & FAIR                 \\
         & 1     & 0.52 & UNFAIR               \\ \hline
\end{tabular}
\end{table}

\subsection{Statistical Parity Difference}
For the completed model we next look at Statistical Parity Difference (SPD).  The closer the result is to 0, the more fair it is.

\begin{equation}
    SPD = P(\hat{Y} = 1 \mid A = 0) - P(\hat{Y} = 1 \mid A = 1),
\end{equation}

where \(\hat{Y}\) is the models predictions, \(A=0\) is the protected attribute for the unprivileged class and \(A=1\) is the protected attribute for the privileged class. \cite{irfan2023}.  The results can be seen for each variable in the table below.

\begin{table}[h]
\centering
\caption{Statistical Parity Difference (SPD) by Variable}
\label{tab:spd}
\begin{tabular}{llcc}
\hline
Variable & Class & SPD & SPD Difference \\ \hline
GENDER   & Male & 0.53 & 0.07 \\
         & Female & 0.46 & \\
AGE      & Under 40 & 0.51 & 0.04 \\
         & 40 Plus & 0.47 & \\
VET      & Veteran & 0.34 & 0.16 \\
         & Non-Veteran & 0.50 & \\
EDUC     & No College & 0.51 & 0.05 \\
         & College & 0.46 & \\
MARSTAT  & Married/Previously Married   & 0.55 & 0.13           \\
         & Never Married                & 0.42 &                \\
EMPLOY   & Unemployed                   & 0.49 & 0.02           \\
         & Employed                     & 0.51 &                \\
RACE     & White                        & 0.51 & 0.03           \\
         & Non-White                    & 0.48 &                \\
ETHNIC   & Hispanic/Latino              & 0.51 & 0.16           \\
         & NonHispanic/Latino           & 0.35 &                \\
PREG     & Pregnant                     & 0.47 & 0.03           \\
         & Not Pregnant                 & 0.50 &                \\ \hline
\end{tabular}
\end{table}

\subsection{Statistical Parity Difference – Multiclass}
This was done again for each class of noprior and the results can be seen in the table below.  The max threshold is the maximum threshold for which a fairness is achieved for the spd value for each class.  If at least one of the classes has an SPD of zero, then SPD is satisfied for that variable, otherwise it is not.
\FloatBarrier
\begin{table}[ht!]
\centering
\caption{Statistical Parity Difference (SPD) Analysis}
\label{tab:spd_analysis}
\resizebox{\columnwidth}{!}{
\begin{tabular}{llccc}
\hline
Variable & Class & SPD  & Max Threshold & At least one SPD = 0 \\ \hline
GENDER   & 0     & 0.27 & 0.26          & Not Satisfied for All Classes \\
         & 3     & 0.07 & 0.06          &                               \\
         & 2     & 0.29 & 0.26          &                               \\
         & 1     & 0.05 & 0.01          &                               \\
AGE      & 0     & 0.17 & 0.16          & Not Satisfied for All Classes \\
         & 3     & 0.30 & 0.26          &                               \\
         & 2     & 0.12 & 0.11          &                               \\
         & 1     & 0.01 & 0.01          &                               \\
VET      & 0     & 0.63 & 0.61          & Satisfied for At Least One Class \\
         & 3     & 0.00 & 0.00          &                               \\
         & 2     & 0.21 & 0.21          &                               \\
         & 1     & 0.17 & 0.16          &                               \\
EDUC     & 0     & 0.05 & 0.01          & Satisfied for At Least One Class \\
         & 3     & 0.00 & 0.00          &                               \\
         & 2     & 0.29 & 0.26          &                               \\
         & 1     & 0.08 & 0.06          &                               \\
MARSTAT  & 0     & 0.14 & 0.11          & Not Satisfied for All Classes \\
         & 3     & 0.18 & 0.16          &                               \\
         & 2     & 0.10 & 0.06          &                               \\
         & 1     & 0.06 & 0.06          &                               \\
EMPLOY   & 0     & 0.42 & 0.41          & Satisfied for At Least One Class \\
         & 3     & 0.00 & 0.00          &                               \\
         & 2     & 0.00 & 0.00          &                               \\
         & 1     & 0.08 & 0.06          &                               \\
RACE     & 0     & 0.02 & 0.01          & Not Satisfied for All Classes \\
         & 3     & 0.35 & 0.31          &                               \\
         & 2     & 0.32 & 0.31          &                               \\
         & 1     & 0.05 & 0.01          &                               \\
ETHNIC   & 0     & 0.14 & 0.11          & Satisfied for At Least One Class \\
         & 3     & 0.00 & 0.00          &                               \\
         & 2     & 0.32 & 0.31          &                               \\
         & 1     & 0.20 & 0.16          &                               \\
PREG     & 0     & 0.58 & 0.56          & Satisfied for At Least One Class \\
         & 3     & 0.21 & 0.21          &                               \\
         & 2     & 0.00 & 0.00          &                               \\
         & 1     & 0.12 & 0.11          &                               \\ \hline
\end{tabular}
}
\end{table}

\FloatBarrier
\subsection{Equal Opportunity}
When the true positive rate (TPR) is the same for both the privileged and unprivileged groups, it is considered fair \cite{irfan2023}.

\begin{equation}
P\left(\hat{Y}=1 \mid Y=1, A=0\right) = P\left(\hat{Y}=1 \mid Y=1, A=1\right)
\end{equation}

\begin{equation}
\text{TPR}_i = \frac{\text{TP}_i}{\text{TP}_i + \text{FN}_i}
\end{equation}

\begin{equation}
\text{EqOpp}_{\text{diff}} = \max_i(\text{TPR}_i) - \min_i(\text{TPR}_i)
\end{equation}

This fairness measure was ran at four different C values with the default values for gamma and r, where \(gamma = scale = 1/n\) and \(r =0\).  C was tried at .1, 1, 10, and 100.  The optimal results are shown in the table below. Notice again that this is tried at different thresholds.  The difference is found and then the maximum threshold value for which the \({EqOpp}_{diff}<threshhold\) is recorded in the table for each model.

\begin{table}[ht!]
\centering
\caption{Equal Opportunity - COMPLETED}
\label{tab:equal_opportunity_completed}
\resizebox{\columnwidth}{!}{
\begin{tabular}{lllllll}
\hline
Variable & Optimal C Value & Model   & Max TPR & Min TPR & EqOpp TPR Diff & Fairness \\ \hline
GENDER   & 0.1             & Linear  & 1       & 0       & 1              & UNFAIR   \\
         & 10              & Poly    & 1       & 0       & 1              & UNFAIR   \\
         & 10              & RBF     & 1       & 0       & 1              & UNFAIR   \\
         & 1               & Sigmoid & 1       & 0       & 1              & UNFAIR   \\
AGE      & 0.1             & Linear  & 1       & 0       & 1              & UNFAIR   \\
         & 10              & Poly    & 1       & 0       & 1              & UNFAIR   \\
         & 10              & RBF     & 1       & 0       & 1              & UNFAIR   \\
         & 1               & Sigmoid & 1       & 0.23    & 0.77           & UNFAIR   \\
VET      & 0.1             & Linear  & 0.99    & 0       & 0.99           & UNFAIR   \\
         & 10              & Poly    & 0.99    & 0       & 0.99           & UNFAIR   \\
         & 10              & RBF     & 0.99    & 0       & 0.99           & UNFAIR   \\
         & 1               & Sigmoid & 1       & 0       & 1              & UNFAIR   \\
EDUC     & 0.1     & Linear  & 0.99    & 0      & 0.99           & UNFAIR   \\
         & 10      & Poly    & 1       & 0      & 1              & UNFAIR   \\
         & 10      & RBF     & 1       & 0      & 1              & UNFAIR   \\
         & 1       & Sigmoid & 1       & 0      & 1              & UNFAIR   \\
MARSTAT  & 0.1     & Linear  & 1       & 0.7    & 0.3            & UNFAIR   \\
         & 10      & Poly    & 1       & 0.7    & 0.3            & UNFAIR   \\
         & 10      & RBF     & 1       & 0.7    & 0.3            & UNFAIR   \\
         & 1       & Sigmoid & 1       & 0.36   & 0.64           & UNFAIR   \\
EMPLOY   & 0.1     & Linear  & 0.99    & 0      & 0.99           & UNFAIR   \\
         & 10      & Poly    & 1       & 0      & 1              & UNFAIR   \\
         & 10      & RBF     & 1       & 0      & 1              & UNFAIR   \\
         & 1       & Sigmoid & 1       & 0      & 1              & UNFAIR   \\
RACE     & 0.1     & Linear  & 1       & 0      & 1              & UNFAIR   \\
         & 10      & Poly    & 1       & 0      & 1              & UNFAIR   \\
         & 10      & RBF     & 1       & 0      & 1              & UNFAIR   \\
         & 1       & Sigmoid & 1       & 0.47   & 0.53           & FAIR     \\
ETHNIC   & 0.1     & Linear  & 1       & 0      & 1              & UNFAIR   \\
         & 10      & Poly    & 1       & 0      & 1              & UNFAIR   \\
         & 10      & RBF     & 1       & 0      & 1              & UNFAIR   \\
         & 1       & Sigmoid & 1       & 0      & 1              & UNFAIR   \\
PREG     & 0.1     & Linear  & 1       & 0      & 1              & UNFAIR   \\
         & 10      & Poly    & 1       & 0      & 1              & UNFAIR   \\
         & 10      & RBF     & 1       & 0      & 1              & UNFAIR   \\
         & 1       & Sigmoid & 1       & 0      & 1              & UNFAIR   \\
\hline
\end{tabular}
}
\end{table}

Next we do the same thing but for each class and take the max difference and the minimum difference to calculate the fairness in each class where c is in each class.  The results can be seen in the table below.  The  TPR difference is shown in the table and this is the maximum value that the threshold can be taken at.  Again this is shown for the optimal C value with the default values of gamma, r, and degree.

\begin{equation}
    TPR_{sc} = \frac{TP_{sc}}{TP_{sc} + FN_{sc}}
\end{equation}

\begin{equation}
    TPR_{\text{diff}} = \max(TPR_{sc}) - \min(TPR_{sc})
\end{equation}

\begin{equation}
    TPR_{sc} = \frac{TP_{sc}}{TP_{sc} + FN_{sc}}
\end{equation}

\begin{equation}
    FPR_{sc} = \frac{FP_{sc}}{FP_{sc} + TN_{sc}}
\end{equation}

\begin{equation}
    \text{Max\_TPR}_{\text{diff}} = \max_c \left( \max_s (TPR_{sc}) \right)
\end{equation}

\begin{equation}
    \text{Min\_TPR}_{\text{diff}} = \min_c \left( \min_s (TPR_{sc}) \right)
\end{equation}

\begin{equation}
    \text{EqOpp}_{\text{diff}} = \left| \text{Max}_{\text{TPR}_{\text{diff}}} - \text{Min}_{\text{TPR}_{\text{diff}}} \right|
\end{equation}

\begin{table*}[ht!]
\centering
\caption{Equalized Opportunity - NOPRIOR}
\label{tab:equalized_opporutnity_noprior}
\begin{tabular}{|c|c|c|c|c|c|c|}
\hline
\textbf{Variable} & \textbf{Optimal C Value} & \textbf{Model} & \textbf{TPR Difference} & \textbf{FPR Difference} & \textbf{Equalized Odds Diff} & \textbf{Fair/Unfair} \\ \hline
Variable & C=Optimal & Model & Max TPR & Min TPR & Equalized Opp Diff & Fair/Unfair \\
\hline
GENDER & 10 & Linear & 0.69 & 0.37 & 0.32 & UNFAIR \\
 & 1 & Poly & 0.70 & 0.56 & 0.14 & FAIR \\
 & 10 & RBF & 0.69 & 0.52 & 0.17 & FAIR \\
 & 1 & Sigmoid & 0.65 & 0.40 & 0.25 & UNFAIR \\
\hline
\hline
AGE & 10 & Linear & 0.69 & 0.43 & 0.26 & UNFAIR \\
 & 1 & Poly & 0.73 & 0.58 & 0.15 & FAIR \\
 & 10 & RBF & 0.71 & 0.54 & 0.17 & FAIR \\
 & 1 & Sigmoid & 0.67 & 0.44 & 0.23 & UNFAIR \\
\hline
 & 10 & Linear & 0.50 & 0.00 & 0.50 & UNFAIR \\
 & 1 & Poly & 0.63 & 0.00 & 0.63 & UNFAIR \\
 & 10 & RBF & 0.50 & 0.00 & 0.50 & UNFAIR \\
 & 1 & Sigmoid & 0.51 & 0.00 & 0.51 & UNFAIR \\
\hline
EDUC & 10 & Linear & 0.55 & 0.23 & 0.32 & UNFAIR \\
 & 1 & Poly & 0.66 & 0.41 & 0.25 & UNFAIR \\
 & 10 & RBF & 0.63 & 0.38 & 0.25 & UNFAIR \\
 & 1 & Sigmoid & 0.54 & 0.32 & 0.22 & UNFAIR \\ 
MARSTAT & 10 & Linear & 0.50 & 0.49 & 0.01 & FAIR \\
 & 1 & Poly & 0.62 & 0.62 & 0.00 & FAIR \\
 & 10 & RBF & 0.60 & 0.56 & 0.04 & FAIR \\
 & 1 & Sigmoid & 0.54 & 0.45 & 0.09 & FAIR \\
\hline
EMPLOY & 10 & Linear & 0.51 & 0.12 & 0.39 & UNFAIR \\
 & 1 & Poly & 0.62 & 0.50 & 0.12 & FAIR \\
 & 10 & RBF & 0.60 & 0.38 & 0.22 & UNFAIR \\
 & 1 & Sigmoid & 0.51 & 0.25 & 0.26 & UNFAIR \\
\hline
RACE & 10 & Linear & 0.50 & 0.48 & 0.02 & FAIR \\
 & 1 & Poly & 0.60 & 0.62 & 0.02 & FAIR \\
 & 10 & RBF & 0.60 & 0.54 & 0.06 & FAIR \\
 & 1 & Sigmoid & 0.54 & 0.49 & 0.05 & FAIR \\
\hline
ETHNIC & 10 & Linear & 0.50 & 0.00 & 0.50 & UNFAIR \\
 & 1 & Poly & 0.63 & 0.00 & 0.63 & UNFAIR \\
 & 10 & RBF & 59.00 & 0.00 & 59.00 & UNFAIR \\
 & 1 & Sigmoid & 0.50 & 0.50 & 0.00 & FAIR \\
\hline
PREG & 10 & Linear & 0.50 & 0.00 & 0.50 & UNFAIR \\
 & 1 & Poly & 0.62 & 0.00 & 0.62 & UNFAIR \\
 & 10 & RBF & 0.59 & 0.00 & 0.59 & UNFAIR \\
 & 1 & Sigmoid & 0.51 & 0.00 & 0.51 & UNFAIR \\
\hline
\end{tabular}
\end{table*}

\FloatBarrier
\subsection{Equalized Odds}
Let \(A=1\) and \(A= 0\) be the privileged and unprivileged groups respectively.  When \(Y=1\), the equations shows the TPR and when \(Y=0\) it shows the FPR.  This shows the ROC where TPR and FPR are equal for the privileged and unprivileged demographics.\\\cite{irfan2023}.

\begin{equation}
    P\left[\hat{Y}=1 \middle| Y=1, A=0\right] = P[\hat{Y}=1|Y=1, A=1]
\end{equation}

The table below shows the TPR and FPR difference as well as the Equalized odds difference and whether it is fair or unfair.  The differences are calculated by subtracting the two values from each class from each other for the TPF and FPR respectively for each variable.  Fairness is determined based on a .2 threshold.  Any threshold could be chosen.  Since equality determines fairness, a threshold closer to zero is wanted.

\begin{table*}[ht!]
\centering
\caption{Equal Odds - COMPLETED}
\label{tab:equal_odds_completed}
\begin{tabular}{|c|c|c|c|c|c|c|}
\hline
\textbf{Variable} & \textbf{Optimal C Value} & \textbf{Model} & \textbf{TPR Difference} & \textbf{FPR Difference} & \textbf{Equalized Odds Diff} & \textbf{Fair/Unfair} \\ \hline
GENDER  & 0.1  & Linear   & 0.32 & 0.03 & 0.29 & UNFAIR \\
        & 10   & Poly     & 0.14 & 0.09 & 0.05 & FAIR   \\
        & 10   & RBF      & 0.17 & 0.06 & 0.11 & FAIR   \\
        & 1    & Sigmoid  & 0.25 & 0.06 & 0.19 & FAIR   \\ \hline
AGE     & 0.1  & Linear   & 0.26 & 0.03 & 0.23 & UNFAIR \\
        & 10   & Poly     & 0.15 & 0.05 & 0.10 & FAIR   \\
        & 10   & RBF      & 0.16 & 0.08 & 0.08 & FAIR   \\
        & 1    & Sigmoid  & 0.22 & 0.01 & 0.21 & UNFAIR \\ \hline
VET     & 0.1  & Linear   & 0.51 & 0.22 & 0.29 & UNFAIR \\
        & 10   & Poly     & 0.63 & 0.18 & 0.45 & UNFAIR \\
        & 10   & RBF      & 0.60 & 0.16 & 0.44 & UNFAIR \\
        & 1    & Sigmoid  & 0.51 & 0.22 & 0.29 & UNFAIR \\ \hline
EDUC    & 0.1  & Linear   & 0.32 & 0.29 & 0.03 & FAIR   \\
        & 10   & Poly     & 0.25 & 0.48 & 0.23 & UNFAIR \\
        & 10   & RBF      & 0.25 & 0.36 & 0.11 & FAIR   \\
        & 1    & Sigmoid  & 0.22 & 0.00 & 0.22 & UNFAIR \\ \hline
MARSTAT & 0.1  & Linear   & 0.02 & 0.03 & 0.01 & FAIR   \\
        & 10   & Poly     & 0.01 & 0.02 & 0.01 & FAIR   \\
        & 10   & RBF      & 0.04 & 0.01 & 0.03 & FAIR   \\
        & 1    & Sigmoid  & 0.09 & 0.02 & 0.07 & FAIR   \\ \hline
EMPLOY  & 0.1  & Linear   & 0.39 & 0.03 & 0.36 & UNFAIR \\
        & 10   & Poly     & 0.12 & 0.04 & 0.08 & FAIR   \\
        & 10   & RBF      & 0.22 & 0.04 & 0.18 & FAIR   \\
        & 1    & Sigmoid  & 0.26 & 0.01 & 0.25 & UNFAIR \\ \hline
RACE    & 0.1  & Linear   & 0.02 & 0.00 & 0.02 & FAIR   \\
        & 10   & Poly     & 0.00 & 0.02 & 0.02 & FAIR   \\
        & 10   & RBF      & 0.06 & 0.03 & 0.03 & FAIR   \\
        & 1    & Sigmoid  & 0.05 & 0.04 & 0.01 & FAIR   \\ \hline
ETHNIC  & 0.1  & Linear   & 0.50 & 0.17 & 0.33 & UNFAIR \\
        & 10   & Poly     & 0.63 & 0.13 & 0.50 & UNFAIR \\
        & 10   & RBF      & 0.59 & 0.16 & 0.43 & UNFAIR \\
        & 1    & Sigmoid  & 0.00 & 0.25 & 0.25 & UNFAIR \\ \hline
PREG    & 0.1  & Linear   & 0.50 & 0.25 & 0.25 & UNFAIR \\
        & 10   & Poly     & 0.62 & 0.19 & 0.43 & UNFAIR \\
        & 10   & RBF      & 0.59 & 0.21 & 0.38 & UNFAIR \\
        & 1    & Sigmoid  & 0.51 & 0.25 & 0.26 & UNFAIR \\ \hline
\end{tabular}
\end{table*}

\FloatBarrier
\subsection{Equalized Odds - Multiclass}
We do the same thing for the multiclass except now we take the maximum and minumum of each class to get the TPR difference and FPR difference for each class where \(s\in sense\ variable\),\ \(c\in class\ label\).

\begin{equation}
TPR_{sc} = \frac{TP_{sc}}{TP_{sc} + FN_{sc}} 
\end{equation}

\begin{equation}
FPR_{sc} = \frac{FP_{sc}}{FP_{sc} + TN_{sc}} 
\end{equation}
    
\begin{equation}
TPR_{diff} = \max(TPR_{sc}) - \min(TPR_{sc}) 
\end{equation}
    
\begin{equation}  
  FPR_{diff} = \max(FPR_{sc}) - \min(FPR_{sc}) 
\end{equation}

\begin{align}
    EqOdds_{diff} = & \left| \max(TPR_{sc}) - \min(TPR_{sc}) \right| \nonumber \\
    & - \left| \max(FPR_{sc}) - \min(FPR_{sc}) \right|
\end{align}

The table with the TPR difference, FPR difference, Equalized Odds difference and whether it is fair or unfair for each variable can be seen below.  Again this was done at a .2 threshold and the default values for gamma, degree, and r with optimal C values for the multiclass noprior model.  Note that closer to zero for an \({EqOdds}_{diff}\) is more fair.

\begin{table*}[ht]
\centering
\caption{Equalized Odds NOPRIOR}
\label{Equalized_Odds_NOPRIOR}
\begin{tabular}{|c|c|c|c|c|c|c|c|}
\hline
\multicolumn{8}{|c|}{\textbf{ETHNIC}} \\ \hline
\textbf{C-Value} & \textbf{Model} & \textbf{Max TPR} & \textbf{Min TPR} & \textbf{Max FPR} & \textbf{Min FPR} & \textbf{EOD Diff} & \textbf{Fairness} \\ \hline
\multicolumn{8}{|c|}{\textbf{GENDER}}\\\hline
10 & Linear   & 1.00 & 0.00 & 0.26 & 0.00 & 0.74 & UNFAIR \\
1  & Poly     & 1.00 & 0.00 & 0.14 & 0.00 & 0.86 & UNFAIR \\
10 & RBF      & 1.00 & 0.00 & 0.12 & 0.00 & 0.88 & UNFAIR \\
1  & Sigmoid  & 1.00 & 0.00 & 0.29 & 0.00 & 0.71 & UNFAIR \\ \hline
\multicolumn{8}{|c|}{\textbf{AGE}}\\\hline
10 & Linear   & 0.99 & 0.00 & 0.20 & 0.00 & 0.79 & UNFAIR \\
1  & Poly     & 1.00 & 0.00 & 0.07 & 0.00 & 0.93 & UNFAIR \\
10 & RBF      & 0.99 & 0.00 & 0.07 & 0.00 & 0.92 & UNFAIR \\
1  & Sigmoid  & 1.00 & 0.27 & 0.28 & 0.00 & 0.45 & UNFAIR \\ \hline
\multicolumn{8}{|c|}{\textbf{VET}}\\\hline
10 & Linear   & 0.99 & 0.00 & 0.17 & 0.00 & 0.82 & UNFAIR \\
1  & Poly     & 0.99 & 0.00 & 0.17 & 0.00 & 0.82 & UNFAIR \\
10 & RBF      & 0.99 & 0.00 & 0.17 & 0.00 & 0.82 & UNFAIR \\
1  & Sigmoid  & 1.00 & 0.00 & 0.22 & 0.00 & 0.78 & UNFAIR \\ \hline
\multicolumn{8}{|c|}{\textbf{EDUC}}\\\hline
10 & Linear   & 0.99 & 0.00 & 0.09 & 0.00 & 0.90 & UNFAIR \\
1  & Poly     & 1.00 & 0.00 & 0.06 & 0.00 & 0.94 & UNFAIR \\
10 & RBF      & 1.00 & 0.00 & 0.07 & 0.00 & 0.93 & UNFAIR \\
1  & Sigmoid  & 1.00 & 0.00 & 0.31 & 0.00 & 0.69 & UNFAIR \\\hline \multicolumn{8}{|c|}{\textbf{MARSTAT}}\\\hline
10 & Linear   & 1.00 & 0.69 & 0.11 & 0.00 & 0.20 & UNFAIR \\
1  & Poly     & 1.00 & 0.70 & 0.11 & 0.00 & 0.19 & FAIR   \\
10 & RBF      & 1.00 & 0.71 & 0.12 & 0.00 & 0.17 & FAIR   \\
1  & Sigmoid  & 1.00 & 0.29 & 0.34 & 0.00 & 0.37 & UNFAIR \\ \hline
\multicolumn{8}{|c|}{\textbf{EMPLOY}}\\\hline
10 & Linear   & 0.99 & 0.00 & 0.25 & 0.00 & 0.74 & UNFAIR \\
1  & Poly     & 1.00 & 0.00 & 0.06 & 0.00 & 0.94 & UNFAIR \\
10 & RBF      & 1.00 & 0.00 & 0.06 & 0.00 & 0.94 & UNFAIR \\
1  & Sigmoid  & 1.00 & 0.00 & 0.89 & 0.00 & 0.11 & FAIR \\ \hline
\multicolumn{8}{|c|}{\textbf{ETHNIC}} \\ \hline
10 & Linear   & 1.00 & 0.00 & 0.09 & 0.00 & 0.91 & UNFAIR \\
1  & Poly     & 1.00 & 0.00 & 0.06 & 0.00 & 0.94 & UNFAIR \\
10 & RBF      & 1.00 & 0.00 & 0.06 & 0.00 & 0.94 & UNFAIR \\
1  & Sigmoid  & 1.00 & 0.34 & 0.28 & 0.00 & 0.38 & UNFAIR \\ \hline
\multicolumn{8}{|c|}{\textbf{ETHNIC}} \\ \hline
10 & Linear   & 1.00 & 0.00 & 0.08 & 0.00 & 0.92 & UNFAIR \\
1  & Poly     & 1.00 & 0.00 & 0.10 & 0.00 & 0.90 & UNFAIR \\
10 & RBF      & 1.00 & 0.00 & 0.07 & 0.00 & 0.93 & UNFAIR \\
1  & Sigmoid  & 1.00 & 0.00 & 1.00 & 0.00 & 0.00 & FAIR   \\ \hline
\multicolumn{8}{|c|}{\textbf{PREG}} \\ \hline
10 & Linear   & 1.00 & 0.00 & 1.00 & 0.00 & 0.00 & FAIR   \\
1  & Poly     & 1.00 & 0.00 & 1.00 & 0.00 & 0.00 & FAIR   \\
10 & RBF      & 1.00 & 0.00 & 1.00 & 0.00 & 0.00 & FAIR   \\
1  & Sigmoid  & 1.00 & 0.00 & 1.00 & 0.00 & 0.00 & FAIR   \\ \hline
\end{tabular}
\end{table*}

\FloatBarrier
\section{The Model}
\subsection{Support Vector Machines}
The data sets are split 70/30 training and test sets with shuffle set at true which splits the data randomly.  Support Vector Machines were run with four kernels.  Linear, Polynomial, Radial Basis Function, and Sigmoid.  The formulas for these kernels can be seen below.

\begin{align}
    \text{Linear Kernel}: & \nonumber \\
    & \hspace{-2cm} K(X,Y) = X^T Y \\
    \text{Polynomial Kernel}: & \nonumber \\
    & \hspace{-2cm} K(X,Y) = (X^T Y + C)^d \\
    \text{Radial Basis Function Kernel}: & \nonumber \\
    & \hspace{-2cm} K(X,Y) = e^{-\gamma \|X - Y\|^2}, \\
    & \hspace{-2cm} \gamma > 0 \nonumber \\
    \text{Sigmoid Kernel}: & \nonumber \\
    & \hspace{-2cm} K(X,Y) = \tanh(\gamma X^T Y + C)
\end{align}
    
\begin{align}
    \text{where } \gamma = \text{auto} \quad & = \frac{1}{n} \\
    \text{where } \gamma = \text{scale} \quad & = \frac{1}{n \times \text{Var}(X)} \\
    \text{where } \text{Var}(x) \quad & = \frac{1}{n} \sum_{i=1}^{n} (x_i - \mu)^2
\end{align}

After further research, it was determined that with algorithms such as SVM, feature importance is not that useful.  It is more useful to select an appropriate C value instead.  The C value is a parameter that sets how much you want to penalize your model for each misclassified point.  This means that the larger the C value, the smaller the margin, the smaller the C value, the larger the margin.  The goal is to maximize the margin while minimizing the classification error.  Next, we looked at different C values.  In the models above the C value was set to four different C-values, .1, 1, 10, and 100 and the optimal values were chosen for each model. Note in the figure below \cite{felipe2019} the difference between large and small C values.
    
\begin{figure}[ht!]
  \includegraphics[width=0.50\textwidth]{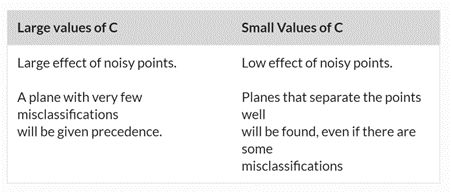}
\caption{The Influence of C Parameter \cite{Ref1}}
\label{fig:13}       
\end{figure}

The results of using the different C-Values with the default values of \textit{gamma}, \textit{r}, and \textit{degree} can be seen on the COMPLETED and NOPRIOR datasets for all four kernels below. The best performing model was the polynomial model with a C value of 10 having a prediction accuracy of \text{63.70\%} for the COMPLETED data set. The confusion matrix can be seen in the figure below. The radial basis function kernel model was a close second with an accuracy of \text{61.73\%} with a \(C=10\). \(C=10\) performed better in all models except in the sigmoid and linear kernel models.

\begin{figure}[ht!]
  \includegraphics[width=0.35\textwidth]{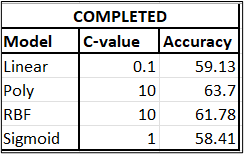}
\caption{Accuracy COMPLETED} 
\label{fig:14}       
\end{figure}

\begin{figure}[ht!]
  \includegraphics[width=0.50\textwidth]{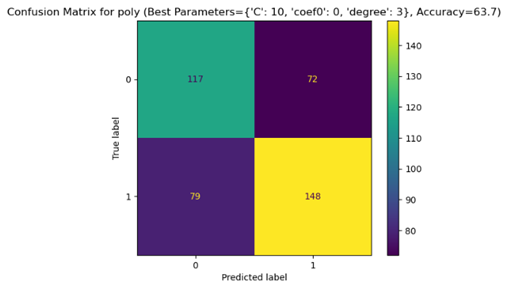}
\caption{Optimal Confusion COMPLETED} 
\label{fig:15}       
\end{figure}

\FloatBarrier
For the multiclass model, the accuracy was much higher with RBF coming in first with an accuracy of \text{91.83\%} with a C-value of 10 and polynomial coming in second with an accuracy of \text{90.89\%} and a C-value of 1.  The linear model was also high at \text{88.09\%} with a C-value of 10.  The confusion matrix can be seen below.

\begin{figure}[ht!]
  \includegraphics[width=0.35\textwidth]{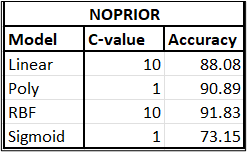}
\caption{Accuracy NOPRIOR} 
\label{fig:16}       
\end{figure}

\FloatBarrier
\begin{figure}[ht!]
  \includegraphics[width=0.50\textwidth]{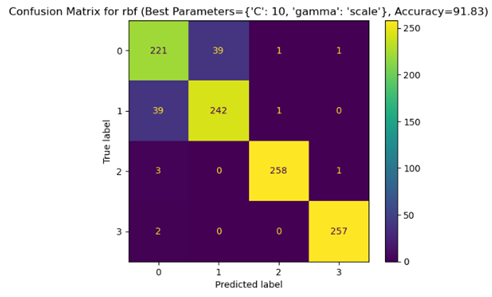}
\caption{Optimal Confusion NOPRIOR} 
\label{fig:17}       
\end{figure}

\FloatBarrier

\subsection{Decision Trees}
A grid search approach was used for decision trees with the following parameters:

\begin{itemize}
    \item \textbf{Max Depth}: [None, 2, 4, 6, 8, 10]
    \item \textbf{Min Samples Split}: [2, 5, 10]
    \item \textbf{Min Samples Leaf}: [1, 2, 4]
\end{itemize}

The optimal results for each data set can be seen below:

\begin{figure}[ht!]
  \includegraphics[width=0.50\textwidth]{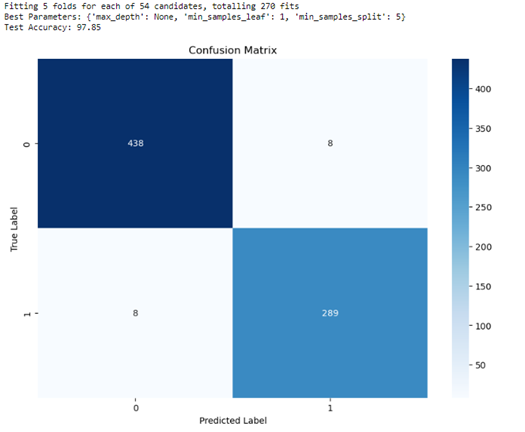}
\caption{Optimal Confusion COMPLETED DT} 
\label{fig:18}       
\end{figure}

\FloatBarrier
\begin{figure}[ht!]
  \includegraphics[width=0.50\textwidth]{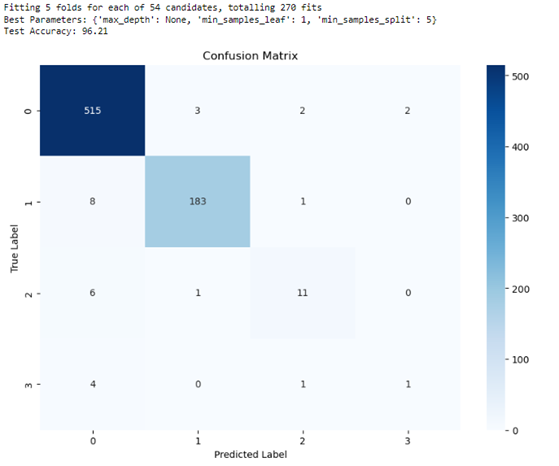}
\caption{Optimal Confusion NOPRIOR DT} 
\label{fig:19}       
\end{figure}

\begin{figure}[ht!]
  \includegraphics[width=0.50\textwidth]{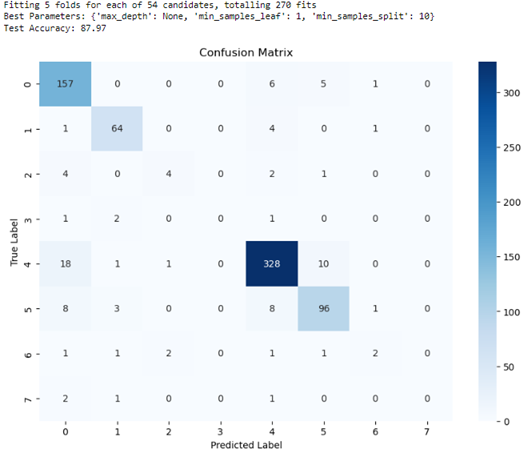}
\caption{Optimal Confusion COMPLETED\_NOPRIOR DT} 
\label{fig:20}       
\end{figure}

\FloatBarrier
\subsection{Random Forests}
A parameter grid approach was used for random forests with the following parameters:

\begin{itemize}
    \item \textbf{Number of Estimators}: [10, 50, 100, 200]
    \item \textbf{Max Features}: ['auto', 'sqrt', 'log2']
    \item \textbf{Max Depth}: [None, 5, 10, 20]
    \item \textbf{Min Samples Split}: [2, 5, 10]
    \item \textbf{Min Samples Leaf}: [1, 2, 4]
\end{itemize}

\begin{figure}[ht!]
  \includegraphics[width=0.50\textwidth]{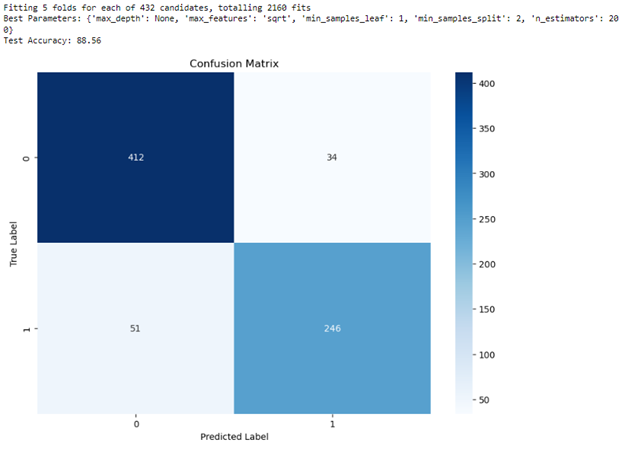}
\caption{Optimal Confusion COMPLETED RF} 
\label{fig:21}       
\end{figure}

\FloatBarrier
\begin{figure}[ht!]
  \includegraphics[width=0.50\textwidth]{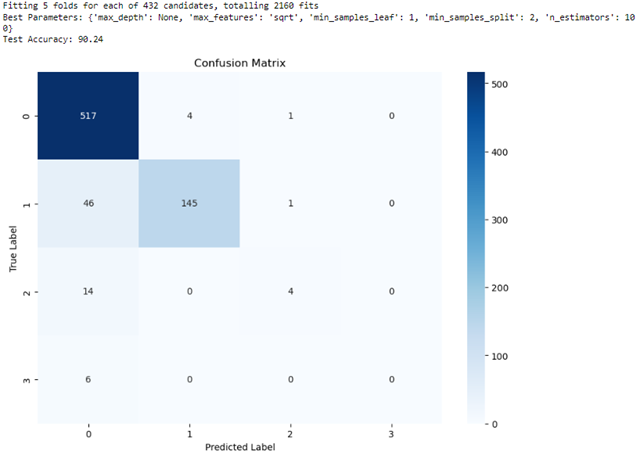}
\caption{Optimal Confusion NOPRIOR RF} 
\label{fig:22}       
\end{figure}

\begin{figure}[ht!]
  \includegraphics[width=0.50\textwidth]{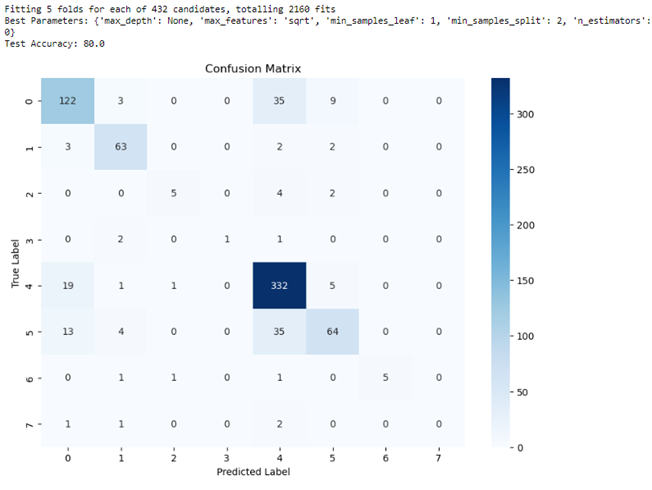}
\caption{Optimal Confusion COMPLETED\_NOPRIOR RF} 
\label{fig:23}       
\end{figure}

\FloatBarrier

\subsection{Neural Networks}
The parameter grid for the neural network model is as follows:

\begin{itemize}
    \item \textbf{Units in Layer 1}: [8, 10, 20, 30]
    \item \textbf{Units in Layer 2}: [8, 10, 20, 30]
    \item \textbf{Activation Function in Input Layers}: ['relu', 'tanh', 'sigmoid']
    \item \textbf{Optimizer}: ['adam', 'sgd']
    \item \textbf{Loss Function}: ['categorical\_crossentropy', 'mean\_squared\_error']
\end{itemize}

The resulting optimal parameters along with their confusion matrices for each data set can be seen below:
\FloatBarrier
\begin{figure}[ht!]
  \includegraphics[width=0.50\textwidth]{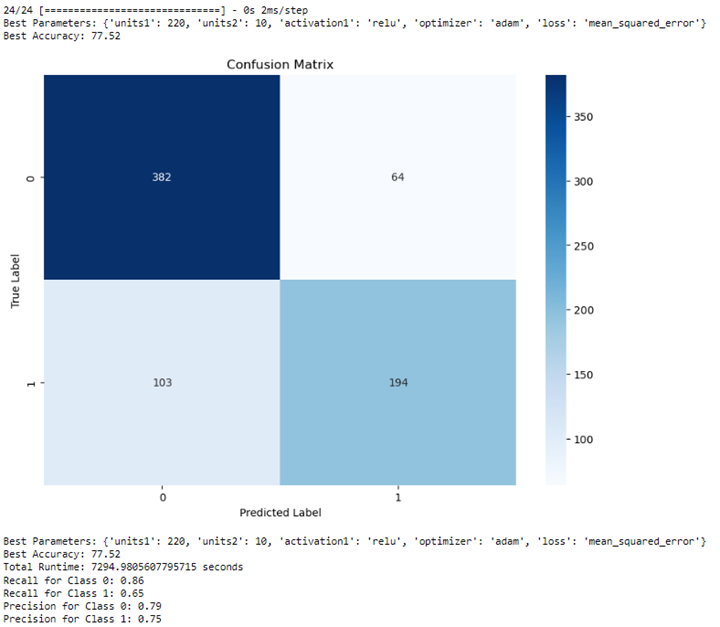}
\caption{Optimal Confusion COMPLETED NN} 
\label{fig:24}       
\end{figure}

\FloatBarrier
\begin{figure}[ht!]
  \includegraphics[width=0.50\textwidth]{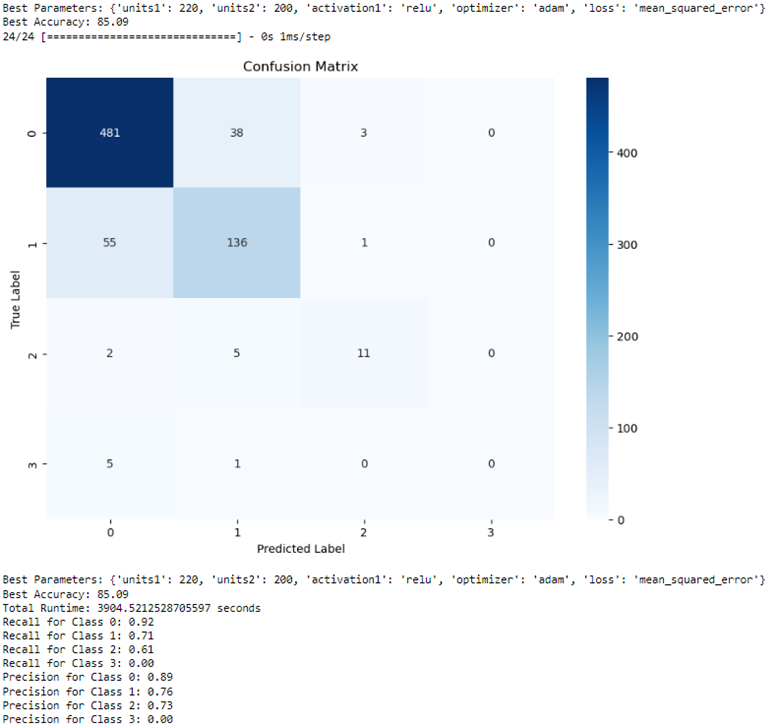}
\caption{Optimal Confusion NOPRIOR NN} 
\label{fig:25}       
\end{figure}

\begin{figure}[ht!]
  \includegraphics[width=0.50\textwidth]{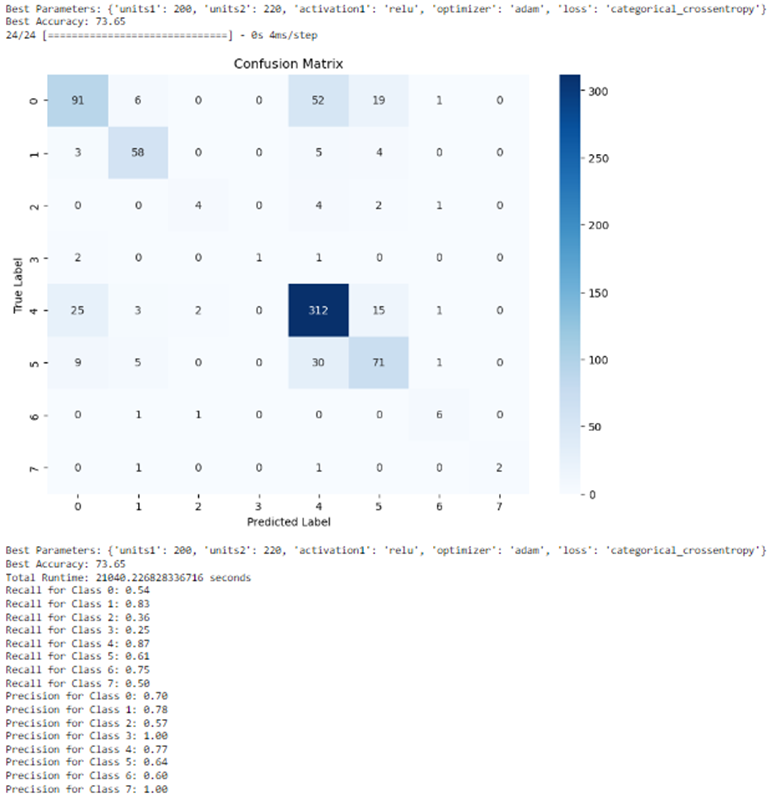}
\caption{Optimal Confusion COMPLETED\_NOPRIOR NN} 
\label{fig:26}       
\end{figure}

\FloatBarrier

\section{Interpretation of Results}

The NOPRIOR variable translation can be seen below.  This is the translation to what is shown in the x-axis of the confusion matrix above. 

\begin{figure}[ht!]
  \includegraphics[width=0.50\textwidth]{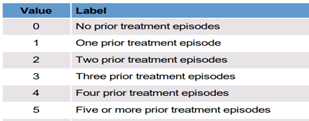}
\caption{NOPRIOR Translation} 
\label{fig:27}       
\end{figure}

The encoded variables used in the data sets and their translated meanings can be seen in the following tables.  

\begin{table}[]
\centering
\begin{tabular}{|p{0.25\linewidth}|p{0.7\linewidth}|}
\hline
\textbf{variable} & \textbf{meaning} \\ \hline
SERVICES\_4 & Rehab/residential, short term (30 days or fewer) \\ 
SERVICES\_5 & Rehab/residential, long term (more than 30 days) \\ 
SERVICES\_7 & Ambulatory, non-intensive outpatient \\ 
PSOURCE\_2 & Alcohol/drug use care provider \\ 
PSOURCE\_3 & Other health care provider \\ 
PSOURCE\_4 & School (educational) \\ 
PSOURCE\_5 & Employer/EAP \\ 
PSOURCE\_6 & Other community referral \\ 
PSOURCE\_7 & Court/criminal justice referral/DUI/DWI \\ 
METHUSE\_2 & No medication-assisted opioid therapy \\ 
LIVARAG\_2 & Dependent living \\ 
LIVARAG\_3 & Independent living \\ 
ALCFLG\_1 & alcohol reported at admissions \\
COKEFLG\_1 & cocaine/crack reported at admissions \\
MARFLG\_1 & marijuana/hashish reported at admissions \\
HERFLG\_1 & heroin reported at admissions \\
METHFLG\_1 & non-rx methadone reported at admissions \\
OPSYNFLG\_1 & Other opiates/synthetics reported at admission \\
PCPFLG\_1 & PCP reported at admission \\
HALLFLG\_1 & Hallucinogens reported at admission \\
MTHAMFLG\_1 & Methamphetamine/speed reported at admission \\
AMPHFLG\_1 & Other amphetamines reported at admission \\
STIMFLG\_1 & Other stimulants reported at admission \\
BENZFLG\_1 & Benzodiazepines reported at admission \\
TRNQFLG\_1 & Other tranquilizers reported at admission \\
BARBFLG\_1 & Barbiturates reported at admission \\
SEDHPFLG\_1 & Other sedatives/hypnotics reported at admission \\
INHFLG\_1 & Inhalants reported at admission \\
OTCFLG\_1 & Over-the-counter medication reported at admission \\
OTHERFLG\_1 & Other drug reported at admission \\
ALCDRUG\_1 & alcohol only \\
ALCDRUG\_2 & other drugs only \\
ALCDRUG\_3 & alcohol and other drugs \\
DETNLF\_3 & retired, disables not in labor force \\
DETNLF\_4 & resident of institution not in labor force \\
DETNLF\_5 & other not in labor force \\
DETCRIM\_2 & Formal adjudication process \\
DETCRIM\_3 & probation/parole \\
DETCRIM\_6 & prison \\
NOPRIOR\_1 & one prior treatment \\
NOPRIOR\_2 & two prior treatment \\
NOPRIOR\_3 & 3 prior treatment \\
PSOURCE\_2 & Alcohol/drug use care provider \\
PSOURCE\_3 & Other health care provider \\
PSOURCE\_6 & Other community referral \\
PSOURCE\_7 & Court/criminal justice referral/DUI/DWI \\
ARRESTS\_1 & once \\
ARRESTS\_2 & two or more times \\
SUB3\_10 & methamphetamine tertiary \\
SUB3\_11 & other amphetamines tertiary \\
SUB3\_12 & other stimulants tertiary \\
SUB3\_13 & benzodiazepines tertiary \\
SUB3\_16 & other sedatives or hypnotics tertiary \\
SUB3\_19 & other drugs tertiary \\
SUB3\_2 & alcohol tertiary \\
SUB3\_3 & cocaine/crack tertiary \\
SUB3\_4 & marijuana/hashish tertiary \\
SUB3\_5 & heroin tertiary \\
SUB3\_6 & non prescription methadone tertiary \\
SUB3\_7 & other opiates and synthetics tertiary \\
SUB3\_8 & PCP tertiary \\
SUB3\_9 & hallucinogens tertiary \\
\end{tabular}
\caption{Variable Descriptions}
\label{tab:Variable_Descriptions}
\end{table}

\vspace{10mm} 

\begin{table}[htbp]
\centering
\begin{tabular}{|p{0.25\linewidth}|p{0.7\linewidth}|}
\hline
\textbf{variable} & \textbf{meaning} \\ \hline
METHUSE\_2 & no medication-assisted opioid therapy \\
PSYPROB\_2 & no coocurring mental \\
&and substance abuse disorders \\
DSMCRIT\_10 & cannabis use \\
DSMCRIT\_11 & other substance use \\
DSMCRIT\_12 & opioid abuse \\
DSMCRIT\_13 & cocaine abuse \\
DSMCRIT\_2 & substance-induced disorder \\
DSMCRIT\_3 & alcohol intoxication \\
DSMCRIT\_4 & alcohol dependency \\
DSMCRIT\_5 & opioid dependency \\
DSMCRIT\_6 & cocaine dependency \\
DSMCRIT\_7 & cannabis dependency \\
DSMCRIT\_8 & other substance dependency \\
DSMCRIT\_9 & alcohol abuse \\
SUB1\_10 & Methamphetamine/speed primary \\
SUB1\_11 & other amphetamines primary \\
SUB1\_12 & other stimulants primary \\
SUB1\_13 & benzodiazepines primary \\
SUB1\_19 & other drugs primary \\
SUB1\_2 & alcohol primary \\
SUB1\_3 & cocaine/crack primary \\
SUB1\_4 & marijuana/hashish primary \\
SUB1\_5 & heroin primary \\
\end{tabular}
\caption{Variable Descriptions Cont'd}
\label{tab:Variable_Descriptions Cont'd}
\end{table}

\FloatBarrier
\section{Reweighting}
\subsection{Interactions}

The Chi Squared test was used to test significant interactions between bias variables.

\begin{equation}
    \chi^2 = \frac{{(E - O)}^2}{E},
\end{equation}

where E is the expected value and O is referring to the observed frequencies in the contingency table respectively.   We looked at both dual and three-way interactions and then combined the datasets and calculated the probability that at least one of these interactions occurred as our new reweighted weight.  If there is a significant interaction, then the interaction counts for each x1, x2 for the dual interaction or x1, x2, and x3 for the three-way interaction are then divided by the total value counts for x1, x2, with the target or x1, x2, and x3 with the target. This gives the probability.  These would be multiplicative models \cite{veenstra2011}.

\subsubsection{Dual Interaction}
\begin{align}
    S_1 &= \left\{x_1, x_2, \text{target}\right\} \nonumber \\
    S_2 &= \left\{x_1, x_2\right\} \nonumber \\
    \text{DualInteraction}_{\text{Probability}} &= \frac{|S_1|}{|S_2|} \\
    \text{Example:} \nonumber \\
    & \hspace{-4em} \frac{P(\text{Complete} \cap \text{White} \cap \text{Male})}{P(\text{White} \cap \text{Male})}
\end{align}

\subsubsection{Three Way Interaction}
\begin{align}
    S_3 &= \left\{x_1, x_2, x_3, \text{target}\right\} \\
    S_4 &= \left\{x_1, x_2, x_3\right\} \\
    \text{3wayInteraction}_{\text{Probability}} &= \frac{|S_3|}{|S_4|} \\
    \text{Example:} & \nonumber \\
    & \hspace{-4em} \frac{P(\text{Complete} \cap \text{White} \cap \text{Female} \cap \text{Pregnant})}{P(\text{White} \cap \text{Female} \cap \text{Pregnant})}
\end{align}

\subsection{Intersectionalization}
By taking the case for which at least one interaction occurs, we are reweighting the fairness.  Then we combined the two-way and three-way interaction datasets together to get our new df dataset.  This allows for the case when a person can be in two groups at once.  An example would be veteran and male or veteran and male under 40.

\begin{align}
    \text{FinalProbability} &= 1 - \left(1 - \prod \text{probability}_{\text{interactions}}\right) \\
    \text{probability}_{\text{interactions}} &= \left(\text{DualInteraction}_{\text{Probability}}\right) \nonumber \\
    &\quad \cup \left(\text{3wayInteraction}_{\text{Probability}}\right) 
\end{align}

\section{New Fairness Calculations}

\subsection{Worst Case Demographic Parity}
Demographic parity compares the pass rate of (positive outcome) of two groups. It is satisfied if:

\begin{equation}
    P\left(\hat{Y} \middle| A \in \text{{sg}}_i\right) = P\left(\hat{Y} \middle| A \in \text{{sg}}_i\right) \forall i, j \in \mathbb{N},\ i \neq j
\end{equation}

Where N is the total number of subgroups.  When using the worst-case ratio formula, it is also known as demographic parity ratio, \cite{ghosh2021}:

\begin{equation}
    \text{DPR} = \frac{\min \left\{ P\left(\hat{Y} \middle| A \in \text{{sg}}_i\right) \forall i \in \mathbb{N} \right\}}{\max \left\{ P\left(\hat{Y} \middle| A \in \text{{sg}}_i\right) \forall i \in \mathbb{N} \right\}}
\end{equation}

\begin{figure}[ht!]
  \includegraphics[width=0.50\textwidth]{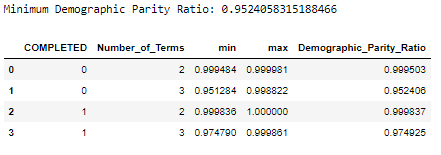}
\caption{Merged - COMPLETED} 
\label{fig:28}       
\end{figure}

\begin{figure}[ht!]
  \includegraphics[width=0.50\textwidth]{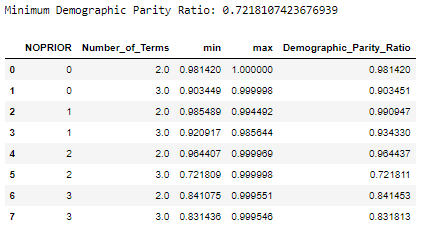}
\caption{Merged - NOPRIOR} 
\label{fig:29}       
\end{figure}

\begin{figure}[ht!]
  \includegraphics[width=0.50\textwidth]{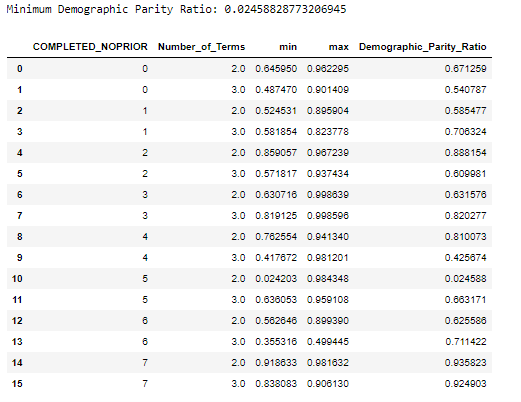}
\caption{Merged - COMPLETED\_NOPRIOR} 
\label{fig:30}       
\end{figure}

\FloatBarrier
\subsection{Worst Case Demographic Parity – Multiclass}
\begin{equation}
    \text{DPR}_{\text{multiclass}} = \min \left\{ \frac{\min \left\{ P\left(\hat{Y} \middle| A \in \text{{sg}}_i\right) \forall i \in N \right\}}{\max \left\{ P\left(\hat{Y} \middle| A \in \text{{sg}}_i\right) \forall i \in N \right\}} \right\}
\end{equation}

\FloatBarrier
The same is done for the multiclass DPR except the overall minimum is taken to get the minimum over all classes. [Gosh et al., 2022]\cite{ghosh2021}.  The results can be seen in the figures below for COMPLETED, NORPIOR, and \text{COMPLETED\_NORPIOR} merged on two-way and three-way interactions.

\begin{figure}[ht!]
  \includegraphics[width=0.50\textwidth]{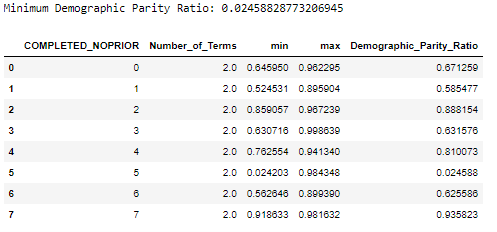}
\caption{2-Way Interaction NOPRIOR} 
\label{fig:31}       
\end{figure}

\begin{figure}[ht!]
  \includegraphics[width=0.50\textwidth]{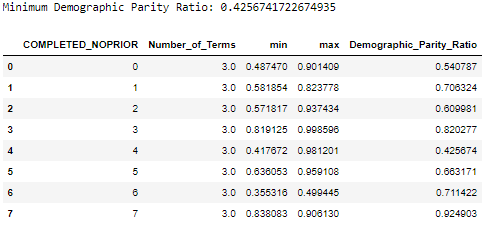}
\caption{3-Way Interaction NOPRIOR} 
\label{fig:32}       
\end{figure}

\subsection{Worst Case Disparate Impact }
\begin{equation}
    \text{DI} = \min \left\{ \frac{P(\hat{Y} | A \in \text{sg}_i)}{P(\hat{Y} | A \in \text{sg}_i)}; \forall i, j \in N, i \neq j \right\},
\end{equation}

where in this case i and j are unique pairs in the series.  The data is grouped by number of terms (2- or 3-way interactions), The probabilities are calculated between the pairs of ratios for each group of target class and pair of probabilities. The minimum ratios are all aggregated across all pairs and then minimum amongst these aggregated ratios is the worst-case scenario for disparate impact. The 80\% rule has to be passed for fairness to be achieved.  This is also known as the four fifths rule.  The pass rate of group 1 to group 2 has to be greater than 80\% with group 1 and 2 being interchangeable \cite{ghosh2021}.  Again, the results are shown below.

\begin{figure}[ht!]
  \includegraphics[width=0.50\textwidth]{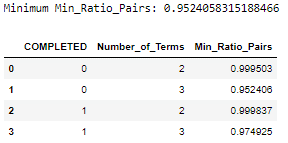}
\caption{Merged COMPLETED DI} 
\label{fig:33}       
\end{figure}

\begin{figure}[ht!]
  \includegraphics[width=0.50\textwidth]{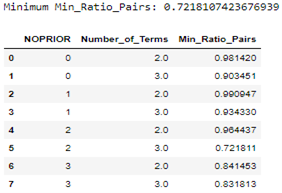}
\caption{Merged NOPRIOR DI} 
\label{fig:34}       
\end{figure}

\begin{figure}[ht!]
  \includegraphics[width=0.50\textwidth]{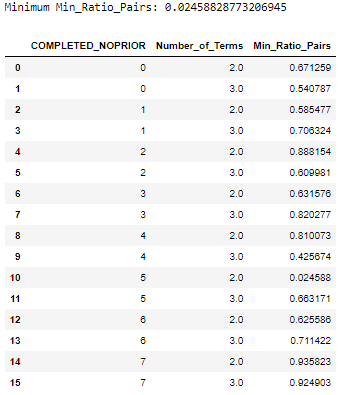}
\caption{Merged COMPLETED\_NOPRIOR DI} 
\label{fig:35}       
\end{figure}

\FloatBarrier
Note that, COMPLETED is the fairest, NOPRIOR is second most fair, and COMPLETED\_NOPRIOR is the least fair.  It is fair individually, however in a worst-case scenario it has the lowest score.

\subsection{Conditional Statistical Parity Ratio}
This uses the worst-case min-max ratio, with predictor \( \hat{Y} \), member \( A \), and legitimate attribute \( L \) \cite{corbettdavies2017}.

\begin{align}
    P\left(\hat{Y} \mid L=1, A \in \text{sg}_i\right) = \nonumber \\P\left(\hat{Y} \mid L=1, A \in \text{sg}_j\right) \quad \forall i, j \in N, i \neq j   
\end{align}

\begin{multline}
\begin{split}
    CSPR &= \\
    \frac{\min\left\{P(\hat{Y} \mid L=1, A \in \text{sg}_i) \ \forall i \in N\right\}}{\max\left\{P(\hat{Y} \mid L=1, A \in \text{sg}_i) \ \forall i \in N\right\}}
\end{split}
\end{multline}

\begin{equation}
\begin{split}
    {CSPR}_{\text{Multiclass}} &= \\
    \min\left\{\frac{\min\left\{P(\hat{Y} \mid L=1, A \in \text{sg}_i) \ \forall i \in N\right\}}{\max\left\{P(\hat{Y} \mid L=1, A \in \text{sg}_i) \ \forall i \in N\right\}}\right\}
\end{split}
\end{equation}

This uses the worst-case min-max ratio, with predictor $\hat{Y}$, member A, and legitimate attribute L \cite{corbettdavies2017}.

\begin{figure}[ht!]
  \includegraphics[width=0.45\textwidth]{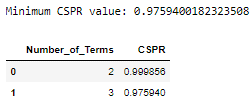}
\caption{CSPR COMPLETED} 
\label{fig:36}       
\end{figure}

\begin{figure}[ht!]
  \includegraphics[width=0.45\textwidth]{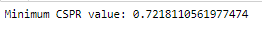}
\caption{CSPR NOPRIOR} 
\label{fig:37}       
\end{figure}

\begin{figure}[ht!]
  \includegraphics[width=0.45\textwidth]{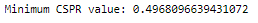}
\caption{CSPR NOPRIOR\_COMPLETED} 
\label{fig:38}       
\end{figure}

\subsection{Equal Opportunity Ratio - Multiclass}
The TPR is compared for the protected and unprotected group.  The minimum of these overall is taken for the multiclass case \cite{ghosh2021}.
\begin{align}
    P(&\hat{Y}=1 | A \in \text{{sg}}_i, Y=1) = \nonumber \\
    &P(\hat{Y}=1 | A \in \text{{sg}}_j, Y=1) \nonumber \\
    &\forall i, j \in N, i \neq j \\
    \text{EOppR} &= \frac{\min \left\{ P(\hat{Y}=1 | A \in \text{{sg}}_i, Y=1) \forall i \in N \right\}}{\max \left\{ P(\hat{Y}=1 | A \in \text{{sg}}_i, Y=1) \forall i \in N \right\}} \\
\end{align}
\begin{equation}
\begin{split}
    \text{EOppR}_{\text{Multi}} &= \\
    \min \left\{ \frac{\min \left\{ P(\hat{Y}=y_k | A \in \text{sg}_i, Y=y_k), \forall i \in N, \forall k \in K \right\}}{\max \left\{ P(\hat{Y}=y_k | A \in \text{sg}_i, Y=y_k), \forall i \in N, \forall k \in K \right\}} \right\}
\end{split}
\end{equation}

\begin{figure}[ht!]
  \includegraphics[width=0.50\textwidth]{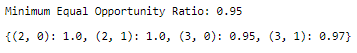}
\caption{Merged COMPLETED Equal Opporunity} 
\label{fig:39}       
\end{figure}

\begin{figure}[ht!]
  \includegraphics[width=0.50\textwidth]{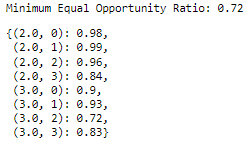}
\caption{Merged NOPRIOR Equal Opportunity} 
\label{fig:37}       
\end{figure}

\begin{figure}[ht!]
  \includegraphics[width=0.50\textwidth]{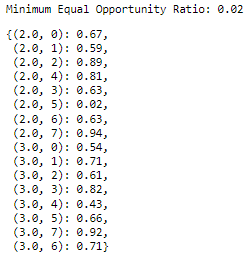}
\caption{Merged COMPLETED\_NOPRIOR Equal Opportunity} 
\label{fig:40}       
\end{figure}

\FloatBarrier
\clearpage
\subsection{Equal Odds Ratio}
K is the set of all possible output classes.  The closer the value is to 1, the lower the disparity is.   For this case we calculate the worst-case odds for each output class \( y \) and then take the minimum of all those values \cite{ghosh2021}. This is the multiclass case for equalized odds ratio.

\begin{align}    
    P\left(\hat{Y}=y_k \middle| A \in \text{{sg}}_i\right) = \nonumber\\P\left(\hat{Y}=y_k \middle| A \in \text{{sg}}_j\right) \quad \forall i, j \in N, i \neq j, \forall k \in K
\end{align}

\begin{align}
    &\text{{EOddR}}_{\text{Multiclass}} = \nonumber\\
    &\hspace{-1.5em}\min\left\{ \frac{\min\left\{P\left(\hat{Y}=y_k \middle| A \in \text{{sg}}_i\right)\right\}}{\max\left\{P\left(\hat{Y}=y_k \middle| A \in \text{{sg}}_i\right)\right\}}\right\} \forall i \in N, \forall k \in K
\end{align}

\begin{figure}[ht!]
  \includegraphics[width=0.50\textwidth]{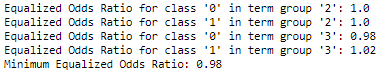}
\caption{Merged COMPLETED Equalized Odds} 
\label{fig:41}       
\end{figure}

\begin{figure}[]
  \includegraphics[width=0.50\textwidth]{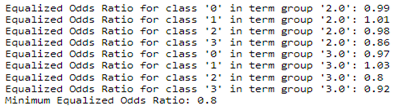}
\caption{Merged NOPRIOR Equalized Odds} 
\label{fig:42}       
\end{figure}

\begin{figure}[]
  \includegraphics[width=0.50\textwidth]{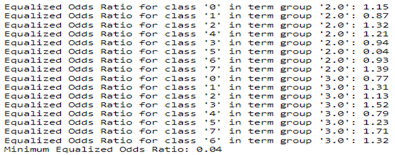}
\caption{Merged COMPLED\_NOPRIOR Equalized Odds} 
\label{fig:43}       
\end{figure}

\begin{figure*}[]
  \centering
  \includegraphics[width=\textwidth]{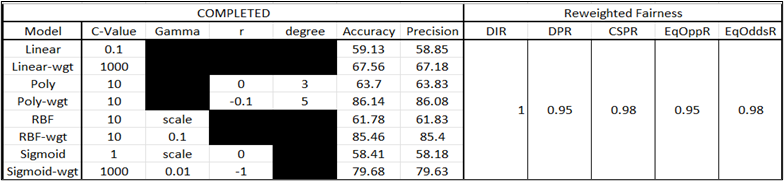}
  \caption{Completed Results} 
  \label{fig:44}       
\end{figure*}

\begin{figure*}[ht!]
  \centering
  \includegraphics[width=\textwidth]{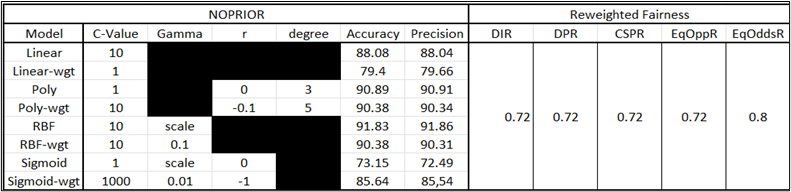}
  \caption{NOPRIOR Results} 
  \label{fig:45}       
\end{figure*}

\begin{figure*}[ht!]
  \centering
  \includegraphics[width=\textwidth]{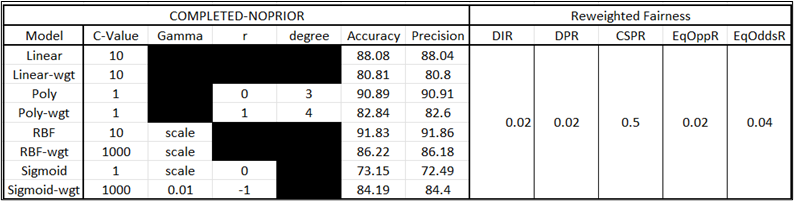}
  \caption{Completed\_NOPRIOR Results} 
  \label{fig:46}       
\end{figure*}

\FloatBarrier

\begin{table}[ht!]
\centering
\begin{tabular}{|c|c|c|c|c|}
\hline
\textbf{Class} & \textbf{Precision} & \textbf{Recall} & \textbf{F1-Score} & \textbf{Support} \\ \hline
Incomplete & 0.98 & 0.99 & 0.98 & 446 \\ \hline
Complete & 0.98 & 0.97 & 0.97 & 297 \\ \hline
\multicolumn{5}{|c|}{\textbf{Accuracy: 97.98}} \\ \hline
\end{tabular}
\caption{Classification Report Completed DT}
\end{table}

\begin{table}[ht!]
\centering
\begin{tabular}{|c|c|c|c|c|}
\hline
\textbf{Class} & \textbf{Precision} & \textbf{Recall} & \textbf{F1-Score} & \textbf{Support} \\ \hline
0 & 0.97 & 0.99 & 0.98 & 522 \\ \hline
1 & 0.97 & 0.95 & 0.96 & 192 \\ \hline
2 & 0.69 & 0.61 & 0.65 & 18 \\ \hline
3 & 0.50 & 0.17 & 0.25 & 6 \\ \hline
\multicolumn{5}{|c|}{\textbf{Accuracy: 96.07}} \\ \hline
\end{tabular}
\caption{Classification Report NOPRIOR DT}
\end{table}

\FloatBarrier
\begin{table}[ht!]
\centering
\centering
\resizebox{\columnwidth}{!}{%
\begin{tabular}{|c|c|c|c|c|}
\hline
\textbf{Class} & \textbf{Precision} & \textbf{Recall} & \textbf{F1-Score} & \textbf{Support} \\ \hline
COMPLETE\_0 & 0.81 & 0.93 & 0.87 & 169 \\ \hline
COMPLETE\_1 & 0.91 & 0.91 & 0.91 & 70 \\ \hline
COMPLETE\_2 & 0.57 & 0.36 & 0.44 & 11 \\ \hline
COMPLETE\_3 & 0.00 & 0.00 & 0.00 & 4 \\ \hline
INCOMPLETE\_0 & 0.94 & 0.91 & 0.93 & 358 \\ \hline
INCOMPLETE\_1 & 0.82 & 0.84 & 0.83 & 116 \\ \hline
INCOMPLETE\_2 & 1.00 & 0.25 & 0.40 & 8 \\ \hline
INCOMPLETE\_3 & 0.00 & 0.00 & 0.00 & 4 \\ \hline
\multicolumn{5}{|c|}{\textbf{Accuracy: 87.97}} \\ \hline
\end{tabular}%
}
\caption{Classification Report Completed\_NOPRIOR DT}
\end{table}

\begin{table}[ht!]
\centering
\begin{tabular}{|c|c|c|c|c|}
\hline
\textbf{Class} & \textbf{Precision} & \textbf{Recall} & \textbf{F1-Score} & \textbf{Support} \\ \hline
Incomplete & 0.89 & 0.92 & 0.91 & 446 \\ \hline
Complete & 0.88 & 0.83 & 0.85 & 297 \\ \hline
\multicolumn{5}{|c|}{\textbf{Accuracy: 88.56}} \\ \hline
\end{tabular}
\caption{Classification Report Completed RF}
\end{table}

\begin{table}[ht!]
\centering
\begin{tabular}{|c|c|c|c|c|}
\hline
\textbf{Class} & \textbf{Precision} & \textbf{Recall} & \textbf{F1-Score} & \textbf{Support} \\ \hline
0 & 0.89 & 0.99 & 0.94 & 522 \\ \hline
1 & 0.97 & 0.76 & 0.85 & 192 \\ \hline
2 & 0.67 & 0.22 & 0.33 & 18 \\ \hline
3 & 0.00 & 0.00 & 0.00 & 6 \\ \hline
\multicolumn{5}{|c|}{\textbf{Accuracy: 90.24}} \\ \hline
\end{tabular}
\caption{Classification Report NOPRIOR RF}
\end{table}
\FloatBarrier

\begin{table} 
\centering
\resizebox{\columnwidth}{!}{%
\begin{tabular}{|c|c|c|c|c|}
\hline
\textbf{Class} & \textbf{Precision} & \textbf{Recall} & \textbf{F1-Score} & \textbf{Support} \\ \hline
COMPLETE\_0 & 0.77 & 0.72 & 0.75 & 169 \\ \hline
COMPLETE\_1 & 0.84 & 0.90 & 0.87 & 70 \\ \hline
COMPLETE\_2 & 0.71 & 0.45 & 0.56 & 11 \\ \hline
COMPLETE\_3 & 1.00 & 0.25 & 0.40 & 4 \\ \hline
INCOMPLETE\_0 & 0.81 & 0.93 & 0.86 & 358 \\ \hline
INCOMPLETE\_1 & 0.78 & 0.55 & 0.65 & 116 \\ \hline
INCOMPLETE\_2 & 1.00 & 0.62 & 0.77 & 8 \\ \hline
INCOMPLETE\_3 & 0.00 & 0.00 & 0.00 & 4 \\ \hline
\multicolumn{5}{|c|}{\textbf{Accuracy: 80.0}} \\ \hline
\end{tabular}%
}
\caption{Classification Report Completed\_NOPRIOR RF}
\end{table}

\clearpage

\begin{table}[htbp]
\centering
\resizebox{\textwidth}{!}{
\begin{tabular}{llll}
\toprule
Model & Completed & NOPRIOR & Completed\_NOPRIOR \\
\midrule
Number of Layers & 2 & 2 & 2 \\
Number of Neurons Layer 1 & 220 & 220 & 200 \\
Number of Neurons Layer 2 & 10 & 200 & 220 \\
Activation Function & ReLU (Rectified Linear Unit) & ReLU (Rectified Linear Unit) & ReLU (Rectified Linear Unit) \\
Network Optimizer & adam & adam & adam \\
Loss Function & mean squared error & mean squared error & categorical crossentropy \\
Epoch & 20 & 20 & 20 \\
Accuracy & 77.52 & 85.09 & 73.65 \\
Recall Incomplete & 0.86 &  &  \\
Recall Complete & 0.65 &  &  \\
Precision Incomplete & 0.79 &  &  \\
Precision Complete & 0.75 &  &  \\
Recall 0 &  & 0.92 &  \\
Recall 1 &  & 0.71 &  \\
Recall 2 &  & 0.61 &  \\
Recall 3 &  & 0.00 &  \\
Precision 0 &  & 0.89 &  \\
Precision 1 &  & 0.76 &  \\
Precision 2 &  & 0.73 &  \\
Precision 3 &  & 0.00 &  \\
Recall 0 &  &  & 0.54 \\
Recall 1 &  &  & 0.83 \\
Recall 2 &  &  & 0.36 \\
Recall 3 &  &  & 0.25 \\
Recall 4 &  &  & 0.87 \\
Recall 5 &  &  & 0.61 \\
Recall 6 &  &  & 0.75 \\
Recall 7 &  &  & 0.50 \\
Precision 0 &  &  & 0.70 \\
Precision 1 &  &  & 0.78 \\
Precision 2 &  &  & 0.57 \\
Precision 3 &  &  & 1.00 \\
Precision 4 &  &  & 0.77 \\
Precision 5 &  &  & 0.64 \\
Precision 6 &  &  & 0.60 \\
Precision 7 &  &  & 1.00 \\
\bottomrule
\end{tabular}
}
\caption{Model performance metrics}
\label{tab:model_performance}
\end{table}

The Rectified Linear Unit activation function is 0 when x is negative and x when x is positive $f(x)=max(x,0)$, \citet{Ramachandran2018}.  This was the optimal activation function in all three models.
\clearpage

\FloatBarrier
\section{Conclusions}
Initially this research set out to predict whether a person completed treatment or not and it proved difficult due to the imbalance of the data with roughly 32\% of people completing treatment. When this proved to have poor accuracy with SVMs, other algorithms such as decision trees and random forest approaches were utilized, A different approach was taken to look at predicting how many prior times a person had been in treatment as well as how the combination of completed and NOPRIOR.  Fairness was explored with 9 variables and after reweighting the variables, a higher fairness was achieved in all the models with only a small decrease in accuracy.  The highest model for the completed data set was the poly-wgt at 86.14\%. Note that the reweighted COMPLETED model is fair when it comes to all fairness measures with a fairness of 95\% for demographic parity, 98\% for CSPR, equal opportunity of 95\% and equal odds of 98\%.  For the NOPRIOR data set, RBF-wgt had the best accuracy with the best fairness.  It had an accuracy of 91.83\% and a fairness of 72\% for all fairness except equal odds which was 80\%.   COMPLETED\_NOPRIOR was fair looking at individual fairness but not at worse case fairness, due to some classes having poor fairness. 
The accuracy of this model was also 91.83\% for RBF-wgt.  Based on fairness and accuracy, the NOPRIOR model predicts the best and COMPLETED is the next best. An RBF or poly kernel would be the best to use when it comes to SVM, c value of 10, r=-0.1, gamma=scale.  This provides the best results.  In comparison, decision trees and random forests performed much better overall with decision trees having an accuracy of 97.98\% for COMPLETED, 96.07\% for NOPRIOR and 87.97\% for COMPLETED\_NOPRIOR.  Random forests had an accuracy of 88.56\% for COMPLETED, 90.24\% for NOPRIOR, and 80\% for COMPLETED\_NOPRIOR.  Neural Networks performed the worst with an accuracy of 77.52\% for COMPLETED, 85.09\% for NORPIOR, and 73.65\% for \newline COMPLETED\_NOPRIOR.  Overall decision trees performed the best with the COMPLETED data set being the best predicting data set, NOPRIOR being next and COMPLETED\_NOPRIOR coming in last in accuracy.  For this reason, we would choose to use the decision trees model to predict all three data sets.

\section{Future Work}
For future work, it would be beneficial to look at linear discriminant analysis for dimensionality reduction.  Also, MCA or Multiple Correspondence Analysis would be good to look at for reducing dimensionality of categorical data.  Researching kernelized MCA would be interesting to see how that affects the accuracy of the model.  Other methods of balancing the data that are more sophisticated than SMOTE using K-nearest neighbors or deep learning methods would be interesting to investigate.  Looking deeper into the data to see what other information could be extracted.  Perhaps a clustering analysis of what drugs/substances occur in treatment together in a patient and how those ties to age and demographics.

\newpage

\bibliographystyle{apalike}
\bibliography{bibliography}

\begin{thebibliography}{}

\bibitem[dsm, 2021]{dsm5}
 (2021).
\newblock {\em Diagnostic and Statistical Manual of Mental Disorders}.
\newblock 5th edition.
\newblock Accessed 21 Dec 2023.

\bibitem[Chawla et~al., 2002]{chawla2002}
Chawla, N.~V., Bowyer, K.~W., Hall, L.~O., and Kegelmeyer, W.~P. (2002).
\newblock Smote: synthetic minority over-sampling technique.
\newblock {\em Journal of Artificial Intelligence Research}, pages 321--357.

\bibitem[Corbett-Davies et~al., 2017]{corbettdavies2017}
Corbett-Davies, S., Pierson, E., Feller, A., Goel, S., and Huq, A. (2017).
\newblock Algorithmic decision making and the cost of fairness.
\newblock In {\em Proceedings of the 23rd ACM SIGKDD International Conference on Knowledge Discovery and Data Mining}, pages 797--806.

\bibitem[Dressel and Farid, 2018]{dressel2018accuracy}
Dressel, J. and Farid, H. (2018).
\newblock The accuracy, fairness, and limits of predicting recidivism.
\newblock {\em Science Advances}, 4(1).

\bibitem[Felipe, 2019]{felipe2019}
Felipe (2019).
\newblock The c parameter.

\bibitem[Gajawada, 2021]{gajawada}
Gajawada, K.~S. (2021).
\newblock Chi-square test for feature selection in machine learning.

\bibitem[Ghosh et~al., 2021]{ghosh2021}
Ghosh, A. et~al. (2021).
\newblock Characterizing intersectional group fairness with worst-case comparisons.
\newblock In {\em AIDBEI}.

\bibitem[Irfan et~al., 2023]{irfan2023}
Irfan, Z., McCaffery, F., and Loughran, R. (2023).
\newblock {\em Evaluating Fairness Metrics}, volume 1840 of {\em Communications in Computer and Information Science}.
\newblock Springer, Cham.

\bibitem[Jishan et~al., 2015]{jishan2015}
Jishan, S., Rashu, R., Haque, N., et~al. (2015).
\newblock Improving accuracy of students’ final grade prediction model using optimal equal width binning and synthetic minority over-sampling technique.
\newblock {\em Decis. Anal.}, 2(1).

\bibitem[{National Survey on Drug Use and Health}, 2022]{nsduh2022}
{National Survey on Drug Use and Health} (2022).
\newblock Key substance use and mental health indicators in the united states: Results from the 2022 national survey on drug use and health.
\newblock Substance Abuse and Mental Health Services Administration.
\newblock Accessed 21 Dec 2023.

\bibitem[Ramachandran et~al., 2018]{Ramachandran2018}
Ramachandran, P., Zoph, B., and Le, Q. (2018).
\newblock Searching for activation functions.

\bibitem[{Substance Abuse and Mental Health Services Administration}, 2021]{samsa2021}
{Substance Abuse and Mental Health Services Administration} (2021).
\newblock Treatment episode data set (teds): 2019.

\bibitem[Veenstra, 2011]{veenstra2011}
Veenstra, G. (2011).
\newblock Race, gender, class, and sexual orientation: intersecting axes of inequality and self-rated health in canada.
\newblock {\em Int J Equity Health}, 10(3).

\bibitem[Wildra and Herring, 2021]{wildra2021}
Wildra, E. and Herring, T. (2021).
\newblock States of incarceration: The global context 2021.
\newblock Prison Policy Initiative.
\newblock Accessed April 21, 2022.

\end{thebibliography}

\end{document}